\documentclass[runningheads]{llncs}
\usepackage{graphicx}
\usepackage{amsmath,amssymb} % define this before the line numbering.
\usepackage{color}
\usepackage[width=122mm,left=12mm,paperwidth=146mm,height=193mm,top=12mm,paperheight=217mm]{geometry}
\usepackage[linesnumbered,ruled,vlined]{algorithm2e}
\usepackage{tabularx}
\usepackage{booktabs}

\usepackage{makecell}
\usepackage{capt-of}

\definecolor{patrick}{rgb}{0.4, 0.2, 0.8}

\definecolor{ben}{rgb}{0.0, 0.58431372549019607843137254901961, 0.65098039215686274509803921568627}

\definecolor{todo}{rgb}{1.0, 0., 0.}

\newcommand{\comment}[1]{}

\begin{document}
% \linenumbers
\pagestyle{headings}
\mainmatter
\def\ECCV18SubNumber{2963}  % Insert your submission number here

\title{DynaMiTe: A Dynamic Local Motion Model with Temporal Constraints for Robust Real-Time Feature Matching} % Replace with your title

\titlerunning{DynaMiTe: A Dynamic Local Motion Model with Temporal Constraints}

\authorrunning{P. Ruhkamp, R. Gong, N. Navab, B. Busam}

\author{Patrick Ruhkamp \quad Ruiqi Gong \quad Nassir Navab \quad Benjamin Busam
{\tt\small p.ruhkamp@tum.de} \quad {\tt\small b.busam@tum.de}
}
\institute{Technische Universit{\"a}t M{\"u}nchen, Germany}

\maketitle

\begin{abstract}
Feature based visual odometry and SLAM methods require accurate and fast correspondence matching between consecutive image frames for precise camera pose estimation in real-time. 
Current feature matching pipelines either rely solely on the descriptive capabilities of the feature extractor or need computationally complex optimization schemes.
We present the lightweight pipeline \textbf{DynaMiTe}, which is agnostic to the descriptor input and leverages spatial-temporal cues with efficient statistical measures.
%, by exploiting the nature of feature point clustering around well defined structures.
The theoretical backbone of the method lies within a probabilistic formulation of feature matching and the respective study of physically motivated constraints.
A dynamically adaptable \textbf{local motion model} encapsulates groups of features in an efficient data structure.
\textbf{Temporal constraints} transfer information of the local motion model across time, thus additionally reducing the search space complexity for matching.
% for finding nearest neighbor descriptors in consecutive frames.
DynaMiTe achieves superior results both in terms of matching accuracy and camera pose estimation with high frame rates, outperforming state-of-the-art matching methods while being computationally more efficient.
%while maintaining real-time applicability on a consumer CPU. We report frame rates of up to \(44fps\) while outperforming other fast methods with GPU acceleration by \(3.5\)-fold.
%\benni{quantify when evaluation is done: Speed comparison number, improvement percentage. Can we say something about complexity? X~ms on CPU?}
%Contrary to the common approach to improve feature matching by increasing the descriptiveness of features, our method shows superior results even with lower descriptive features and is generic to the given explicit kind of features for input. Based on the motion smoothness assumption of neighboring points in the scene, the TP rate of feature correspondences increases with the number of input features. Imposed by temporal constraints the complexity for initial feature matching of \(O\left(n^{2}\right)\) is reduced significantly through minimizing \(n\) by preconditioning matching from the gain of temporal information. Our method outperforms matching accuracy and runtime of other state-of-the-art methods. We show increased performance on pose estimation and also compare frame-to-frame pose estimation for long sequences to fully mature SLAM pipelines.
\end{abstract}

\section{Introduction}\noindent
%\patrick{
%Mention shortcomings of other pipelines! also e.g. GMS and TILDE etc. Story telling: GMS has shown compelling results for wide baseline matching. However, their method assumes certain statistics on the feature point distribution which is not certainly true in all scenarios. Additionally do their assumptions only consider... Therefore, we propose Union-Find-Disjoint-Set (UFDS) for finding local feature clusters around well defined structures in the scene, reliable for accurate camera pose estimation. To further improve the feature matching accuracy, we introduce novel temporal constraints, thus also improving overall runtime for the complete feature matching pipeline even without GPU utilization.}
\textbf{Visual self-localization} from consecutive video frames of a freely moving camera has a long history~\cite{matthies1988incremental} and is one of the key challenges in 3D computer vision.
SLAM methods have been applied in robotics and UAVs~\cite{von2017monocular} and are a crucial element in augmented reality pipelines~\cite{Engel-et-al-pami2018} as well as medical applications~\cite{busam_miccai2018}.
Besides well known % visual odometry and SLAM m
methods based on direct image alignment~\cite{Engel-et-al-pami2018,Engel_ECCV_2014,hanna1991direct,newcombe2011kinectfusion}, different sparse feature based methods are also well studied~\cite{klein07parallel,Mur-Artal2017,klein2006full,Davison2007}.
\begin{figure}[t!]
\centering
\includegraphics[width=1.0\linewidth]{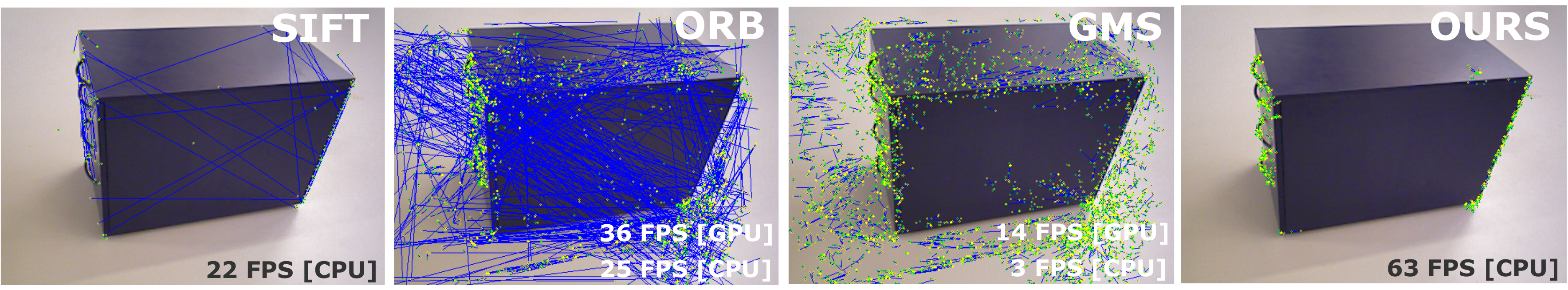}
\caption{Comparison of feature matching for consecutive image frames on a challenging low-textured object of the TUM RGB-D dataset~\cite{sturm2012benchmark}. Features in frame $I_{i}$ in yellow; $I_{j}$ in green; Matches as blue lines. SIFT~\cite{Lowe2004} is texture-sensitive. ORB~\cite{ORB_features} (2000 extractions) is efficient but unstable. GMS~\cite{bian2017gms} produces noisy, wrong matches in uniform regions while our method runs the fastest with minimal incorrect matches.}
\label{Fig:image_teaser}
\end{figure}
%%%\vspace{-1.0em}

\textbf{Direct methods} incorporate the image information directly from pixel intensities, which can be error-prone due to illumination changes, moving objects or shutter effects~\cite{Engel_ECCV_2014}. However, a dense image alignment can help with dense reconstructions of the scene~\cite{cremers2017direct}.
\textbf{Feature based methods} rely on distinctive feature points extracted from the image input, which can account for illumination changes while reducing the computational complexity.
% for image alignment.
Due to their sparseness, they are more suitable for SLAM methods with loop closures and bundle adjustment; however reconstructions are not dense~\cite{Mur-Artal2017}.

The first step in feature based visual odometry and SLAM systems is to detect and to match keypoints between consecutive frames.
Quality and robustness of this step is vital for camera pose estimation and all subsequent computations in the pipeline. 
Errors in pose estimation are usually treated in a second stage by pose optimization with local and global bundle adjustment or graph based optimization schemes~\cite{g2oKuemmerle2011,Strasdat2011}.\\
%This step is computationally expensive, and well estimated initial poses accelerate runtime capabilities and diminish pose drift.
%To avoid the need for frequent large trajectory corrections, we propose a method which is both computationally tractable and delivers highly accurate poses.
%%\vspace{-1.0em}

\textbf{Motivation. }\noindent
\textit{Natura non facit saltus}.\footnote{Latin for "nature does not make jumps".} This principle of natural philosophy was a crucial element in the formulation of infinitesimal calculus and classical mechanics~\cite{baumgarten2013metaphysics}.
%The latter can be described as closed systems of finite-order differential equations, thus yielding differentiable forces.
%Newton's second law of motion $F\left( t \right) = m \cdot a \left( t \right)$ relates the force on an object with its mass and acceleration and thus lays the foundation of motion.
Consequently, as
\begin{equation}
    x\left( t + \Delta t \right)
    %= \iint a \,d^2t
    %= x_0 + v_0\,t + \frac{1}{2}\, a\,t^2,
    \approx x\left( t \right) +  v_0 \Delta t + \frac{1}{2}\, a\,\left(\Delta t\right)^2,
\end{equation}
we assume smooth motion of an object in space, which is also true for its projection $P\left( x \right)$ onto a camera image. Knowledge of the motion at time $t$ thus helps to approximate the projected location $P\left( x + \Delta t \right)$ in the next frame.

\begin{figure}[h]
\centering
\includegraphics[width=0.6\linewidth]{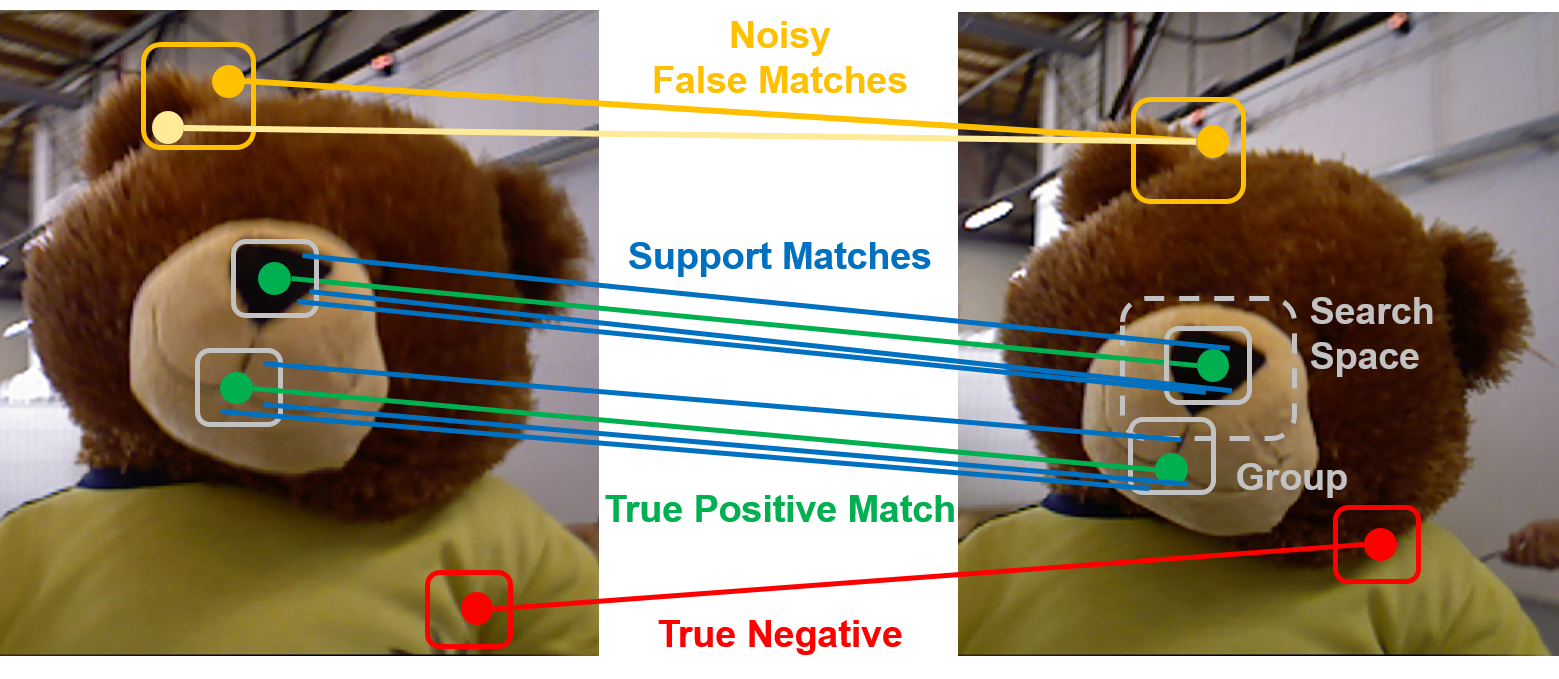}
\caption{Matching and reinforcement. Support matches (blue) between groups of feature points reinforce each other. Areas with little structure or blurry parts can lead to noisy false matches (orange). The proposed locally adaptive clustering algorithm encapsulates proximate feature points. All clusters within a defined search space (dashed line) are potential matching candidates.}
\label{Fig:motivation}
%%%\vspace{-1.0em}
\end{figure}

Given a video sequence, extracted feature points around descriptive parts of the image (e.g. some object in the scene) are grouped into local feature groups by our novel clustering algorithm.
The spatial 2D displacement of corresponding groups is then propagated from previous frames by a motion proxy to constrain the search space for new feature matches.
%Based on the motion proxy for neighboring feature points within some local vicinity of a potential feature match, other matches within this region reinforce the potential match as support matches.
Since close features likely belong to the same scene structure, their motion is similar and inter-frame matches between corresponding groups can reinforce each other.
This is justified by statistical measures based on a binomial distribution detailed in section~\ref{section:probs}. 
It follows that for a certain number of $n$ features in a group, a minimum number of $N$ matches between the groups is needed to confirm a true positive match (cf. Fig.~\ref{Fig:motivation}).\\

%%\vspace{-1.0em}

\textbf{Contributions and Outline. }\noindent
\textit{DynaMiTe} combines two complementary elements of spatially coherent motion and temporarily smooth inter-frame displacements - analogous to its eponym - in its joint formulation for feature matching between consecutive image frames.
%Our fast and robust method merges the two components  into a coupled model, capable of real-time performance.
To this end, \textit{DynaMiTe} contributes:
\begin{enumerate}%[itemsep=0mm]
%%%%\vspace{-0.2em}
\item A \textbf{dynamic local motion model} encapsulating the differentiable spatial motion prior with \textbf{temporal coherency constraints} through frame-to-frame information passing.
\item A \textbf{statistical quality criteria} to determine noise-free feature correspondences.
\item An \textbf{efficient} \textbf{clustering scheme} through a light data structure to form groups of close-by feature points.
\item An \textbf{efficient and robust feature matching pipeline} for camera pose estimation in image sequences that significantly \textbf{improves} the state-of-the-art evaluated on the three datasets KITTI~\cite{geiger2013vision}, TUM RGB-D~\cite{sturm2012benchmark}, and TILDE webcam~\cite{TILDE}.
%    %%%\vspace{-0.2em}
\end{enumerate}
To the best of our knowledge, \textit{DynaMiTe} is the first method that uses a generic data structure to form clusters of feature points and combines spatial and temporal constraints for feature matching, formulated in a unified probabilistic model.
%\todo{OUTLINE!!}
We motivate our method by analyzing the shortcomings of similar approaches in Sec.~\ref{PS}. We then give an overview of the general procedure of \textit{DynaMiTe}, introduce our proposed dynamic local motion model (Sec.~\ref{dllm}), extend the probabilistic model of reinforcing support matches between groups of features (Sec.~\ref{section:probs},~\ref{noise}), and deduce robust statistics from it (Sec.~\ref{statistics}). In the experiments we show matching quality, robustness and repeatability on different datasets for \textit{DynaMiTe} as well as runtime performance, outperforming SOTA even in challenging scenes. 
\section{Related Work}\noindent
%%%\vspace{-3.0em}
\textbf{Feature based }%\noindent
visual odometry methods have shown to achieve the tight real-time constraints to compute accurate camera poses and sparse 3D maps of the scene~\cite{klein07parallel}, even for long sequences~\cite{Mur-Artal2017}.
Accurate feature matching 
%is one of the key components of such methods which has an 
has immediate effect on the subsequent tasks of pose estimation and map generation~\cite{Nister2004,Hesch2011,Lepetit2009,Penate-Sanchez2013,Xiao-ShanGao2003,fathian2017quaternion}.
Pose interpolation~\cite{busam2016,shoemake1985} and filtering~\cite{busam2018} techniques can be utilized to circumvent the real-time constraint for camera pose estimation from video sequences to some extent. 
%Matched features can be used for \textbf{camera pose} estimation via the 5-point algorithm~\cite{Nister2004} and epipolar geometry~\cite{Hesch2011,Lepetit2009,Penate-Sanchez2013,Xiao-ShanGao2003}.
To improve feature matching capabilities, multiple feature \textbf{detectors and descriptors} have been developed~\cite{Lowe2004}, also specifically targeting real-time applications~\cite{ORB_features}. One major area of research focuses on the development of robust descriptors which are less variant and more distinctive, thus enabling better matching performance~\cite{csurka2018handcrafted}. Different descriptors~\cite{Bay2008,Morel2009} and learning based pipelines~\cite{balntas2016learning,tian2017l2,mishchuk2017working,tian2019sosnet} enable a variety of vision applications~\cite{choy_NIPS2016,Simo-Serra2015,TILDE,Yi2016}.
Some scholars design descriptor and detector together~\cite{Yi2016,ono2018lf,SuperPoint,shen2019rf,dusmanu2019d2},
%However, their high frame rate performance can solely be achieved with small images.
%and their keypoint locations are only retrieved with pixel accuracy.
%%%\vspace{-1.0em}
or additionally learn the matching task~\cite{sarlin2019superglue} and also including semantic information~\cite{6136}.
Chli and Davison~\cite{ActiveMatching} propose to actively search for features by propagating information from the previous frame.
Targeting specifically wide baseline, Yu et al.~\cite{yi2018learning} proposed an efficient end-to-end pipeline for learning to find correspondences.
%\todo{Deep Semantic Feature Matching, CVPR 2017}
%Fathian et al.~\cite{fathian2017quaternion} propose an adapted pose estimation method with quaternions, achieving better results compared to the classical 5-point and 8-point algorithm~\cite{Hartley1997}.

%For low frame rate cameras and slow pose estimation algorithms in-between poses are efficiently interpolated, resulting in a smooth camera trajectory.
%\ben{This is an application motivation and can be a little shrunk textwise}
%%%\vspace{-1.0em}

Differentiating between \textbf{true correspondences and mismatches} still remains as primary difficulty.
Methods like the ratio test~\cite{Lowe2004} improve feature matching quality by comparing the best and second best potential feature match.
%within the complete set of feature matches. 
Cross check is an alternative to the ratio test, where the nearest neighbor matches are checked for consistency.
%for one feature point is only considered to be true if the
%nearest neighbor for the matched feature candidate 
%match is also consistent in return.\\
%%%\vspace{-1.0em}
\textbf{Statistical approaches} such as RANSAC~\cite{RANSAC} and its modifications~\cite{PROSAC,LMEDS} are effective to remove outliers but may increase runtime due to their iterative execution, especially for large inputs.
FLANN~\cite{Muja2009FastConfiguration} finds approximate nearest neighbors in large datasets and can improve computation times.% However, accuracy for non-deterministic methods is not guaranteed.
%can be problematic, and building up the search tree for every frame is not efficient.

%As their network utilizes pixel coordinates from other feature detection methods, they are agnostic to the specific feature descriptor used for finding keypoint locations during training.
%%%\vspace{-1.0em}

By grouping \textbf{joint motion pairs}~\cite{Motion_Coherence} different methods have been proposed in order to distinguish between true and false matches~\cite{CODE,Lin2014}. Despite showing compelling results, their elaborate formulations result in complex and costly constraints.
Other methods assume similar motion smoothness by matching patches between images~\cite{ConnellyBarnes,nrdc},
%\todo{PatchMatch: \cite{ConnellyBarnes}}
or learn to match patches~\cite{MatchNet}.
%\todo{MatchNet: Unifying Feature and Metric Learning for Patch-Based Matching, CVPR 2015 (Patch Matching https://www.cv-foundation.org/openaccess/content_cvpr_2015/papers/Han_MatchNet_Unifying_Feature_2015_CVPR_paper.pdf; https://github.com/hanxf/matchnet)}
%\todo{ActiveMatching Active Matching (Margarita Chli and Andrew J. Davison, "Active Matching", in ECCV 2008) https://www.doc.ic.ac.uk/~ajd/Publications/chli_davison_eccv2008.pdf}
Sparse~\cite{Lucas1981a} and dense~\cite{Farneback2003} optical flow algorithms~\cite{Field} also assume neighboring points in 3D to move coherently.

Bian et al.~\cite{bian2017gms} (GMS) were the first to formulate the idea of motion smoothness in space within a \textbf{probabilistic model} utilizing a predefined fixed pixel grid.
%After initial brute force nearest neighbor matching (\(O(n^2)\)) of a high number of feature points ($n \in \left[10^3, 10^5\right] $) with GPU acceleration, false matches are identified based on their location in a predefined fixed pixel grid.
Without GPU acceleration, their method is limited by its initial brute force matching to find potential candidates, many of which are being discarded as mismatches afterwards.
%GMS has shown compelling results for wide baseline matching. However, their method assumes certain statistics on the feature point distribution. 
%%%\vspace{-1.0em}
Ma et al.~\cite{Ma2018} transfer the idea of \textbf{close-by feature point} matching directly to the Euclidean distance within consecutive frames. This approximation does not hold true in general and fails in practice for forward/backward translations, where the depth dependent projection scales non-uniformly.
Also~\cite{wang2018} employ locality information to filter match outliers and Zheng et al.~\cite{zheng2018} compute cluster centers from fixed grid patches to compare between frames. Wrong matches in the grid cells, however, shift the cluster center and the method requires initial brute force matching.
%\ben{the above looked initially very fragmented. I tried to group a little better, but it still is problematic. I would not differentiate between classical / deep methods as hereafter. Try to put them into the above sections.}
%\textbf{Deep Learning Methods. }
%Abgrenzung zu DeepLearning methods

We also focus on improving matching quality by using close-by features for reinforcement, but propose a different clustering scheme, together with an improved probabilistic model for noise-free robust feature matches in video sequences.
Unlike matching patches, we match single features where features around some landmark support each other.
%together with adaptive local clustering and motion coherency throughout video sequences, for robust camera pose tracking. 
%\ben{Make sure to emphasize in which category we fall: feature matching with motion cherence, local clustering and }
%%%\vspace{-0.5em}
\section{Methodology}\noindent
\textbf{Problem Statement.} Recent feature matching approaches~\cite{bian2017gms,zheng2018}
for wide-baseline scenarios 
have introduced a simple \textbf{probabilistic model} to distinguish between correct matches and mismatches, where additional matches of proximate features reinforce each other. 
\label{PS}
They are limited by analyzing those matches on regular grids or require expensive clustering algorithms.
In the former scenario~\cite{bian2017gms}, high quantities of uniformly distributed feature points across the entire image are matched, and supporting matches within a regular grid are analyzed.
%for these derivations to hold true. 
A high number of feature points and uniform sampling lead to pairs of many keypoints with poor descriptor quality, resulting in noisy matches and a heavy computation. 
%Regular grid patches would further restrict the model, especially for cases of rotation and forward/backward translation.
Proposed clustering algorithms as in~\cite{zheng2018} are very restrictive and show large variation based on their input, caused by unstable feature point detection between frames.
%\ben{Tried to remove the judgments above to make it more objective.}
The tight realtime constraint is problematic in both cases, as extracting and matching around 1E5 keypoints~\cite{bian2017gms} is solely possible with GPU acceleration. Expensive clustering algorithms~\cite{zheng2018} aggravate the issue.\\

\textbf{DynaMiTe.} We take inspiration of supporting neighbouring matches~\cite{bian2017gms} and extend the approach with our dynamic local clustering method to form groups of close-by feature points around descriptive landmarks.
\begin{figure*}[!b]
%%%\vspace{-1.0em}
\centering
  \includegraphics[width=1.0\textwidth]{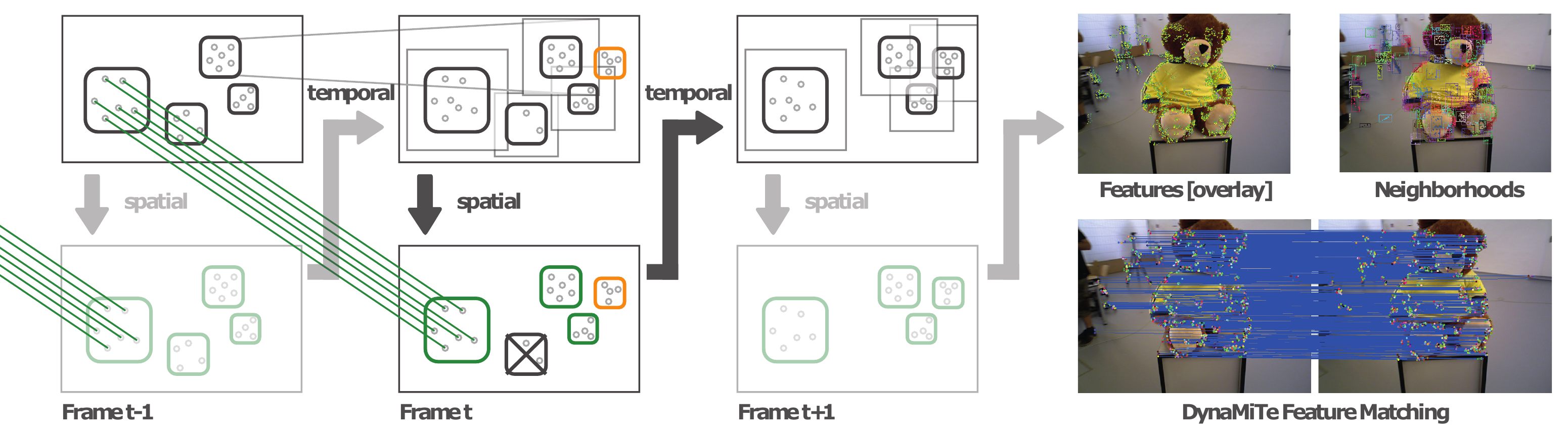}
  \caption{[Left] Schematic illustration of \textit{DynaMiTe} pipeline: Temporal information is passed through the image sequence for each group (upper row). 
  %The search space for potential group associations is enforced by the \textit{temporal constraint} 
  The boxes (light grey) illustrate the enlarged search space around groups between time \(t-1\) and \(t\). 
  %Matches between associated groups are subsequently evaluated by the \textit{spatial constraint}: 
  A feature match is considered true, if enough other matches between the groups can support the match (green).
  %(see matches indicated by green correspondence lines for one exemplary group between \(t-1\) and \(t\)). 
  Groups may disappear (crossed out group at \(t\)) and new ones emerge (orange group at \(t\)). [right] Matched features and groups as overlay on the source image.
  %Extracted groups and matches between subsequent frames are illustrated for \textit{DynaMiTe}.
 }
  \label{pipeline}
  %%\vspace{-1.0em}
\end{figure*}
%into the matching pipeline.
%Additionally we make use of cross check matching (Sec.~\ref{noise}), to further improve robustness and reduce noise. 
After feature computation, 
our proposed method encapsulates the spatial group displacement by the cluster representative.
The \textit{spatial} cluster information is passed throughout the sequence in the \textit{temporal} domain as motion proxy, resulting in a dynamic local motion model. % (see Sec.~\ref{dllm}). % for probabilistic feature matching.
% which is further regularized by \textbf{motion constraints} (cf. Sec.~\ref{sec:temporal_constraint}) 
%embedded in a \textbf{spatial-temporal} formulation
Assuming mainly static scenes and smooth camera motion, only features of clusters within a certain search space around the cluster center in the previous frame need to be considered for matching.
Hence, the group motion is used as prior to restrict the search space for potential matches, which are finally evaluated with our improved and robust probabilistic model.
Algorithm~\ref{DynaMiTe_Overview} gives a general overview of our proposed pipeline, which is schematically detailed in Fig.~\ref{pipeline}.
\begin{center}
\scalebox{0.9}{ 
\centering
\begin{algorithm}[H]
%\KwData{Image \(I_t\); groups \(N_{t-1}\) from \(I_{t-1}\)}
%\KwResult{True group Matches \(\in N_t\)}
\ForAll{Frame $I_{t}$ in video}{
Extract feature points\;
Establish feature groups (FG)\;
Calculate intersection of old and new FGs\;
Match intersected FGs\;
Compute match score and retrieve inlier\;
Establish new FGs\;
Pass FG information to next frame $I_{t+1}$\;
}
\caption{DynaMiTe Pipeline for Feature Matching}
\label{DynaMiTe_Overview}
\end{algorithm}
}
\end{center}

%In the following sections the principles for establishing groups and finding the intersection between them will be detailed.
%In Sec.~\ref{statistics}
We detail the foundation on how to establish a statistical measure for feature matching between patches with a matching score, and improve the base model with bi-directional matching to filter low confidence matches.
%which was not feasible previously due to restrictions of the regular grid model as well as performance limitations of the naive approach.
Additionally, we extend the model with a locally adaptive clustering approach and adapt the underlying statistics for the probabilistic model.
We show that the final probabilistic measure for feature matching only depends on the number of neighbouring feature points within a group and the number of supporting matches.% and thus can be modeled as simple binomial distribution.
%A probabilistic model is derived for computing the match score.%\footnote{We invite the reader to study~\cite{bian2017gms}, as they lay out the fundamentals of the probabilistic model used in Sec.~\ref{section:probs} and Sec.~\ref{statistics}}

\subsection{Dynamic Local Motion Model}
\label{dllm}\noindent
%For now, we expect features between two images already to be matched, such that the probabilistic model from above can be applied as a second filtering step.
%It is unclear how to group and define feature points in the image into clusters. A naive approach would be to assume a regular grid.
%This would mean to expect uniformly distributed features in the image, resulting in a huge number of extracted feature points.
%, which again has severe impact on the initial matching stage.
%Regular grid patches would further restrict the model, especially for cases of rotation and forward/backward translation. Traditional clustering methods may be too restrictive.
%Hence, we propose a fast and dynamic feature grouping approach, by exploiting the nature of many feature detection operators to form clusters around landmarks in the scene.
%%%\vspace{-1.0em}
%\textbf{Dynamic feature grouping with UFDS.} 
We propose a fast and dynamic feature clustering approach by exploiting the nature of many feature detection operators to form clusters around descriptive landmarks in the scene.
For this, \textit{Union-Find Disjoint Sets} (UFDS)~\cite{Galler1964} is utilized for efficient grouping of close-by feature points. The data structure models groups in \textit{DynaMiTe} as collection of disjoint sets.

\begin{figure}[h]
\centering
\includegraphics[width=1.0\linewidth]{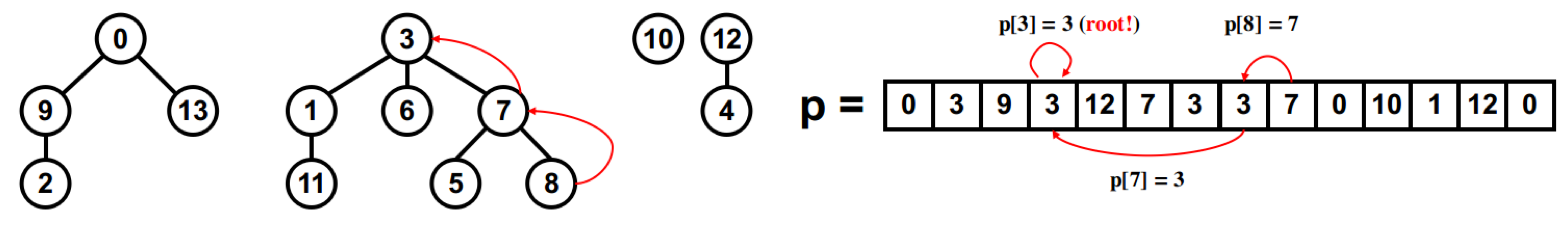}
\caption{Overview of UFDS for efficient clustering of feature points.}
\label{Fig:ufds}
\vspace{-3.0em}
\end{figure}

UFDS is essentially a forest of multi-way trees, where each tree represents a disjoint subset of elements. 
A forest of trees can be implemented as an array $p$ of size $N$ items. $p[i]$ records the index of the parent of item $i$. If $p[i] = i$, then item $i$ is the root of this tree and also the representative item of the subset that contains item $i$ (cf. Fig.~\ref{Fig:ufds}).

This allows to determine which set an item belongs to, check if two items belong to the same set, and merge two disjoint sets into one in nearly constant time (e.g. $O(1)$).
In our 2 dimensional implementation, items are feature points and sets are groups.
%More specifically, the mapping ought to be a data structure by which: 1. given a keypoint, the neighbourhood it belongs to can be identified and 2. given a neighbourhood, all keypoints inside it can be efficiently accessed.
The efficiency of this operation is crucial as identifying the group of keypoints is a frequent operation and the correctness of every match is examined by the correlation between two groups.

Our 2D UFDS data structure considers the maximum size of a cluster in pixels as well as the min. and max. amount of features per group. This is justified by the probabilistic model derived hereafter (cf.~\ref{section:probs}). The analytic matching probabilities give the interval $[5,35]$ as a quality criterion for our group sizes which we also use in all our experiments.
%as depicted in Fig.~\ref{Fig:ProbPlot}
Cluster centers are initialized at random over the set of all extracted feature points. Algorithm details can be found in the suppl. material.
%\vspace{-2.0em}
\subsection{Probabilistic Model}\noindent
\label{section:probs}\noindent
%Generally, we note the probability of a single independent feature to match correctly as \(p(f_{A})=t\).
Similar to~\cite{bian2017gms} we assume that features within a close vicinity will match with a high probability to the same area in another image from a different viewpoint, 
matches of close-by features can reinforce each other.
%To identify a true match, we take some patch to summarize the feature point locations on the image and derive a probabilistic measure on how many feature matches are needed within the patch to support the match.
%need to be established for those matches within the grid cell to identify a true match.
%or patch for being deemed to be true matches.
%In essence, we do not observe individual features, but analyze how many matches can be found between patches. Or figuratively speaking: the more matches between two patches, the better.
%Two scenarios can be distinguished for deriving the probabilistic model: 
After feature points have been grouped with our proposed clustering algorithm, 
we analyze all enclosed features per intersecting groups between video frames.
%a patch which is from the same scene location in consecutive video frames.
%Now, the nearest neighbor for some feature in a patch in the first image should be found in the patch in the other image, observing the same location in the scene. The intuition behind our model is, the more neighboring features are matched between those patches (also noisy false positive matches may contribute), the higher the probability is, that we actually found the true nearest neighbor (NN). 
%Opposed to that, given different patch locations, the number of supporting matches will be small.
The rate of feature matches between patches compared to the number of enclosed keypoints gives a measure of certainty for the match.
%: the higher the ratio the more certain the feature match becomes.
We can derive a probabilistic model by examining the matching events between correlated and uncorrelated image patches and deduce a binomial distribution which is only dependent on the number of enclosed keypoints in the patch. More specifically, we can define a threshold for a true positive match as the minimum amount of supporting feature matches between two groups relative to their enclosed keypoints.

%Examining the possible events as explained in the following, we can derive a binomial distribution for the true positive and true negative cases, which are widely separated and only dependent on the number of feature points within the patch. 
%between those patches needed to verify correct matches between them.
\begin{table}
  \begin{minipage}{0.6\textwidth}
  \centering
  \scalebox{1.0}{
  \begin{tabular}{ll}
Notation & Description \\ \toprule
\(f_{A}\)        &   \makecell[l]{A feature $f$ in $A$ matches correctly; \\ \(p(f_{A})=t\) }    \\
\(\overline{f_{A}}\) & \makecell[l]{A feature $f$ in $A$ matches incorrectly; \\  \(p\left(\overline{f_{A}}\right) = 1-t\)} \\ 
\(T\) & \makecell[l]{Patch $A$ and $B$ view \\ the identical location} \\ 
\(F\) & \makecell[l]{Patch $A$ and $B$ view \\ a different location} \\ 
\(f_{A}^{B}\)      &   NN of $A$ is in $B$   \\
\(\overline{f_{A}^{B}}\)       &   NN of $A$ is NOT in $B$   \\
\midrule
\(p\left(f_{A}, f_{A}^{B}\right)\)      &   \makecell[l]{Probability of \(f\) in $A$ matches correctly \\ AND NN of \(f\) is in $B$}    \\
\(p\left(\overline{f_{A}}, f_{A}^{B}\right)\)      &   \makecell[l]{Probability of \(f\) in $A$ matches wrongly \\ AND NN of \(f\) is in $B$}    \\
\(p\left(f_{A}\mid T\right)\) & \makecell[l]{Probability of \(f\) in $A$ matches correctly \\ GIVEN $T$} \\
\bottomrule
\end{tabular}
}
  \caption{Overview of used notation. NN = Nearest Neighbor in feature space}
  \label{table:notation}
  \end{minipage}
  \hfill
  \begin{minipage}{0.35\textwidth}
  \centering
    \includegraphics[width=0.8\textwidth]{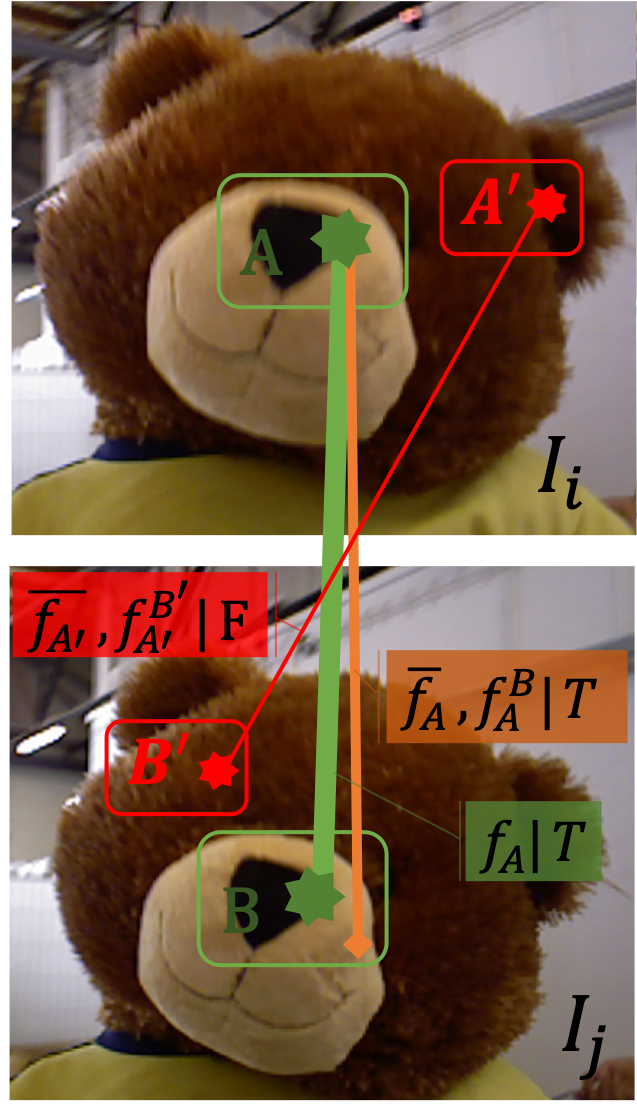}
        \captionof{figure}{Illustration of possible events during feature matching. See text for description and Tab.~\ref{table:notation} for notation.}
    \label{Fig:Events}
\end{minipage}
\vspace{-2.0em}
\end{table}

%Given a patch $A$ in image $I_{i}$ and patch $B$ in image $I_{j}$, each with a certain number of feature points ($f_{i}$,$f_{j}$), we can define the minimal number of matches $m_{i}^{j}$ as our matching criterion.
Figure~\ref{Fig:Events} illustrates the possible matching events (see Table~\ref{table:notation} for notation).
%\ben{Can we bring the table next to the figure? For instance directly underneath, minipage?. Otherwise this is really cumbersome. Also cite in the fig. caption that this is shown in Tab.}
In case of observing corresponding patches \(A,B\) (green case \(T\)) in two images \(I_{i}\) and \(I_{j}\), we can observe a feature (green star) in patch \(A\) that has its nearest neighbor (NN) in descriptor space in patch \(B\) (\(f_{A}^{B}\)). 
This feature can either be correctly matched (\(f_{A}\)), or mismatched with some other feature in \(B\) while its true NN lies still in \(B\) (\(\overline{f_{A}}, f_{A}^{B}\)).
We observe that those mismatches (\(\overline{f_{A}}, f_{A}^{B}\)) still contribute as "noisy" support match between the patches.
%A feature in patch \(A\) may also be mismatched with a feature outside of \(B\), thus not contributing as support match \(\overline{f_{A}},  \overline{f_{A}^{B^{*}}}\).
Similar observations can also be made for the false case \(F\), in which we analyze uncorrelated patches (e.g. patch $A'$ and $B'$ are not identical regions in the scene), where the feature is mismatched to its NN in \(B'\) (\(\overline{f_{A'}}, f_{A'}^{B'}\)).
%We would expect the event of wrongly matching a feature between uncorrelated patches \(\overline{f_{A}},  f_{A}^{B^{\prime}}\) to be small.
%%%\vspace{-1.0em}
By analyzing the matching events, it becomes apparent that there is a high probability of finding multiple matches between correlated patches which support each other. 

\textbf{Mathematical Justification.} 
\label{Justification}
%With the aforementioned observations we can formulate the probabilities to find correct feature matches and mismatches. 
Let $f$ be one of $n$ features in $A$, which we denote to correctly match to some feature out of $N$ features in $B$ 
%(where for now $n<N$ as our pipeline tries to find matches with all intersecting groups within a search space) 
as \(f_{A}\) with \(p(f_{A})=t\).
%For the derivations below 
In case that feature $f$ matches wrongly (i.e. \(\overline{f_{A}}\)), its NN can be any of the other $N$ features in B. % in $B$ assuming we observe the same location, 
Thus, we can write
\begin{equation}
\label{pt}
p\left(f_{A}^{B}\mid \overline{f_{A}}\right) = \frac{n}{N}.
\end{equation}

For correlated patches (case \(T\)), we denote the probability that a feature in $A$ has its NN in $B$ by \(p_{t} = p\left (f_{A}^{B}\mid T\right)\). 
Examining the possible cases for \(p_{t}\) as depicted in Fig.~\ref{Fig:Events}, this consists of a correct match \(p\left(f_{A}\mid T\right)\), or a mismatch while the NN is still in patch $B$ \(p\left(\overline{f_{A}}, f_{A}^{B}\mid T\right)\).
%Let \(p_{t} = p\left (f_{A}^{B}\mid T\right)\) be the probability of a features NN in $A$ is in $B$ given the patches observe the same scene-location, 
Therefore we can write:
\begin{equation}
\label{T1}
\begin{split}
p_{t} = p\left (f_{A}^{B}\mid T\right) &= p\left(f_{A}\mid T\right)+ p\left(\overline{f_{A}},  f_{A}^{B}\mid T\right)\\
&= p\left(f_{A}\mid T\right)+ p\left(\overline{f_{A}}\mid T\right) \cdot p\left(f_{A}^{B}\mid \overline{f_{A}},  T\right).
\end{split}
\end{equation}
With the assumption of independence for single feature matches, we are independent of \(T\).
Using Baye's rule, the notation from Table~\ref{table:notation} and with Eq.~\ref{pt}, we get:
\begin{equation}
\label{T2}
\begin{split}
p_{t} &= p\left (f_{A} \right )+ p\left(\overline{f_{A}}\right )\cdot p\left(f_{A}^{B}\mid \overline{f_{A}}  \right) = t + \left ( 1-t \right ) \cdot \dfrac{n}{N}.
\end{split}
\end{equation}
%The probability for mismatching with the proposed method, can be formulated analogously as:
We assume that each group can be treated equally and that groups have similar numbers of features \(N\).
Analogously for uncorrelated patches \(A'\) and \(B'\) (case \(F\)) we can derive:
%\ben{we assume that each patch can be treated equally. Debatable, but ok. We additionally assume that the amount of features in B' is also N. Also the matching probability is the same. This may be not valid. We should at least mention these two things, right?}
\begin{equation}
\label{F1}
\begin{split}
p_{f} = p\left(f_{A'}^{B'}\mid F  \right ) &=
p\left(\overline{f_{A'}},  f_{A'}^{B'}\mid F  \right ) = p\left(\overline{f_{A'}}\mid F\right )\cdot p\left(f_{A'}^{B'}\mid \overline{f_{A'}},  F  \right)\\
&=p\left(\overline{f_{A'}}\right )\cdot p\left(f_{A'}^{B'}\mid \overline{f_{A'}}  \right) =\left ( 1-t \right ) \cdot \dfrac{n}{N}.
\end{split}
\end{equation}

\subsection{False Positive Reduction}
\label{noise}
\begin{figure}
\vspace{-1.0em}
\centering
\includegraphics[width=0.7\linewidth]{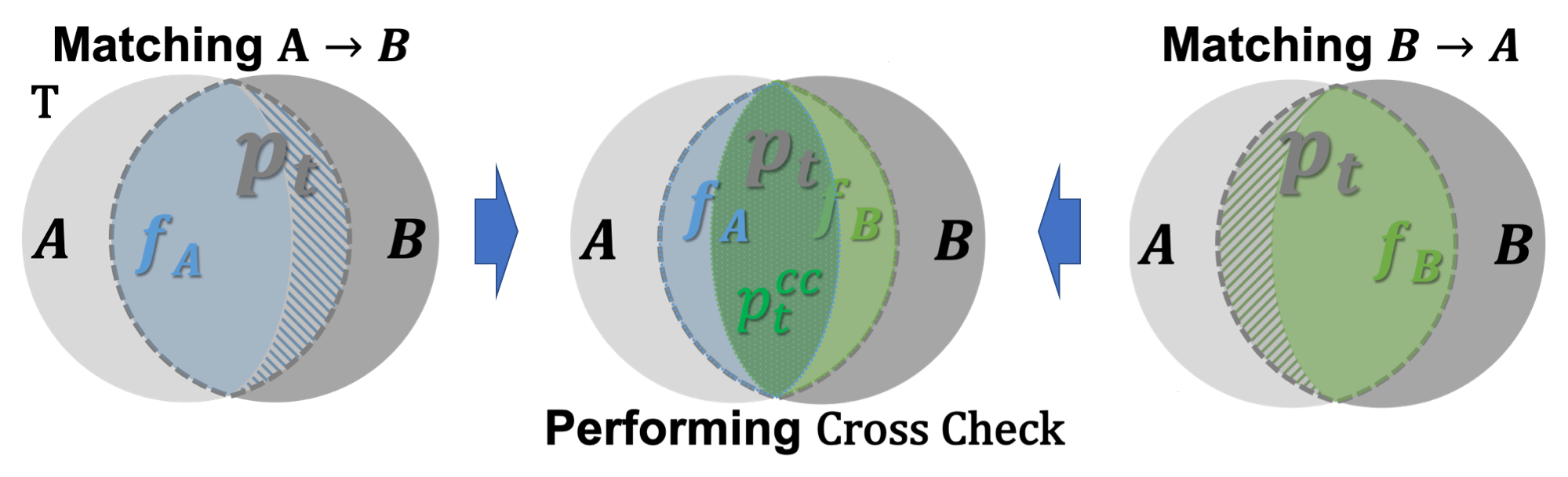}
\caption{Cross check matching for improved robustness and reduced noise. For normal matching ($A$ to $B$ or $B$ to $A$), false matches between patches would still contribute to the matching probability $p_{t}$ as false positives (cf. hatched areas). Cross check consistency results in the union $p_{t}^{cc}$. Area size does not depict probability.}
\label{Fig:Probs}
\vspace{-1.0em}
\end{figure}

\noindent
%It would be desirable to remove false matches between associated patches that have been identified as partners up to now (cf. event \(\overline{f_{A}},  f_{A}^{B}\) in Fig.~\ref{Fig:Events}).
%These mismatches contribute to noisy final matches within the patches (cf. Fig.~\ref{Fig:image_teaser}).
Assuming some feature matches correctly or incorrectly with the same chances, i.e. \(t=0.5\), and with \(n<<N\), we get a wide separation between \(p_{t}\) and \(p_{f}\) (see Eqs.~\eqref{T2} and~\eqref{F1}).
%assuming average matcher with t=0.5 and n<<N (e.g. we do not include any information from the previous frames)... then there is a wide seperation between true and false matches. 
However, this is partly due to including noisy false positive matches, which is not desirable (compare noise for GMS in Fig.~\ref{Fig:image_teaser}).
%Reducing these noisy matches would not affect the separation from above significantly in Eq.~\ref{T2}:
%\(p_{t}\) becomes \(t\) without the small additive factor.
To reduce noisy false positive matches (e.g. event (\(\overline{f_{A}},  f_{A}^{B}\)) in Fig.~\ref{Fig:Events}), we introduce a consistency check via bidirectional matching (compare Fig.~\ref{Fig:Probs}).
However, bidirectional matching has an influence on the terms in Eq.~\ref{T2}.
Details on the derivation below are given in the suppl. material.
%However, this results in smaller probabilities as illustrated in Fig.~\ref{Fig:Probs}.

\textbf{True Matches.} Given correct patch associations \(T\), cross check \(cc\) helps to reduce noisy matches.
%as \(p_{t}^{cc} = p\left (f_{A}^{B}\mid T, cc\right)\). 
%Examining the set of events for \(p_{t}^{cc}\), this consists of either a correct nearest neighbor \(p\left(f_{A}\mid T\right)\), an incorrect one despite observing the same group in the image \(p\left(f_{A},f_{A}^{B}\mid T\right)\), or 
%All nearest neighbors which have been discarded by the cross check \( p\left(\text{Cross-Check false}\right)\). 
%As we perform cross check, we only receive a correct match if the nearest neighbor estimation is also consistent.
%(see equations~\eqref{TMP1} and~\eqref{TMP2} \benni{why the second?}).
%\todo{find names for the formulas such that the author can more easily relate to them}
%Let \(p_{cct} = p\left (f_{a}^{b}t\mid T^{ab},f_{cc}  \right )\) be the probability that, given we consider the same image regions in two views and performing cross check for brute force matching, a feature \(f_{a}\)'s nearest neighbor is in \(b\) and vice versa while also being a true match during cross check.
%\todo{what is cross check}
Let, similar to Eq.~\eqref{T1}, the probability of a feature in \(A\) having its NN in \(B\) under cross check be \(p_{t}^{cc} = p\left (f_{A}^{B}\mid T, cc\right)\), then it holds:
\begin{equation}
\label{TMP1}
\begin{split}
p_{t}^{cc} &= p\left (f_{A}^{B}\mid T,cc  \right ) \\
&= \left(p\left(f_{A}\mid T\right) + p\left(\overline{f_{A}},  f_{A}^{B}\mid T\right)\right) \cdot \left(p\left(f_{B}\mid T\right) +  p\left(\overline{f_{B}},f_{B}^{A}\mid T  \right)\right).% + p\left(cc\text{ false}\right),
\end{split}
\end{equation}
%where $p\left(cc\text{ false}\right) \approx 0$ as a feature that matches from $A$ to $B$, but not vice versa will be discarded.\\
As before (see Eqs.~\eqref{T1} and~\eqref{T2}), with \(m\) and \(M\) being the equivalent for \(n\) and \(N\) and by substitution after the binomial expansion, we can reduce this to:
\begin{equation}
\label{eq:cross_check_prob}
%\begin{split}
p_{t}^{cc} = 
%&p\left (f_{A} \right )\cdot p\left(f_{B} \right ) + p\left(\overline{f_{A}}\right )\cdot p\left(f_{A}^{B}\mid \overline{f_{A}}  \right) \cdot p\left(\overline{f_{B}}\right )\cdot p\left(f_{B}^{A}\mid \overline{f_{B}}  \right)\\
t^{2} + 2 \cdot t \cdot \left ( 1-t \right )\dfrac{n}{N} + \left ( 1-t \right )^{2} \dfrac{n}{N} \cdot \dfrac{m}{M}%\\
%&\approx t^{2} + \left ( 1-t \right )^{2} \cdot \dfrac{n^{2}}{N^{2}}.
%\end{split}
\end{equation}
%with \(n\), \(m\) being the matched feature points in $A$ and \(N\), \(M\) matched feature points in group \(B\).
%%\vspace{-1.0em}
%\ben{We did not introduce M and m.}

\textbf{False Matches.} In analogy for uncorrelated patches it holds:
\begin{equation}
\label{FMP}
\begin{split}
p_{f}^{cc} &= p\left(\overline{f_{A'}},  f_{A'}^{B'}\mid F  \right )\cdot p\left(\overline{f_{B'}},  f_{B'}^{A}\mid F  \right) %+p\left(cc \text{ false}\right)\\
%&=p\left(\overline{f_{A'}}\right )\cdot p\left(f_{A'}^{B'}\mid \overline{f_{A'}}  \right) \cdot p\left(\overline{f_{B'}}\right )\cdot p\left(f_{B'}^{A'}\mid \overline{f_{B'}}  \right)\\
=\left ( 1-t \right )^{2} \dfrac{n}{N} \cdot \dfrac{m}{M}%\\
%&\approx \left ( 1-t \right )^{2} \cdot \dfrac{n^{2}}{N^{2}}.
\end{split}
\end{equation}

\subsection{Robust Statistics}
\label{statistics}
%%%\vspace{-1.0em}
%Following previous derivations, we are now able to separate between true and false matches with a high certainty.
%However, introducing cross check comes at the cost of lowering the probability to find a true match.
%Given our \textbf{dynamic local motion model} and the proposed \textbf{motion constraints} we are able to increase $t$ in our formulation.
Naive bidirectional matching between all features is expensive, especially for the extraction of a large quantity (around $1E5$) of uniformly distributed features in the image as in~\cite{bian2017gms}.
%, where the initial feature extraction and matching is already the bottleneck, and the statistics are applied as a subsequent step.
Additionally, the fraction of \(n/N\) becomes small, as a few features \(n\) in a patch are compared against all features \(N\) of the entire image, which would reduce the separation between \(p_{t}\) and \(p_{f}\). 
%Both effects are not desired.
%, even though it would not degrade the model, as \(p_{f}\) goes to \(0\).

As our proposed model embeds spatial and temporal information and can serve as a motion proxy of the displacement of encapsulated feature points, the potential feature matches are restricted to the intersecting clusters within a certain search space.
Thus, not only the computational bottleneck is reduced, but also \(N\) decreases significantly.
%The last step is due to the fact that we match consecutive frames of a video sequence where \(n\approx N\) (matching $A$ to $B$) and \(m\approx M\) (matching $B$ to $A$).
With the assumption of small inter-frame motion, the number of features in $A$ and $B$ are similar ($n\approx N$) and the fraction in Eq.~\ref{eq:cross_check_prob} and Eq.~\ref{FMP} reaches \(1\), yielding again a wide separation between \(p_{t}\) and \(p_{f}\). Additionally, we suppose $p\left(f_{A}\right)=t$ to be larger than $0.5$ which increases the wide seperation.

\textbf{Matching Quality Criterion.} Matching of an individual feature is generally independent of other features. Thus, we can use the derivations from above similar to~\cite{bian2017gms} to formulate a binomial distribution which describes the probability of finding additional support matches between correlated or uncorrelated groups for some feature match \(m_{i}^{j}\).
Our matching quality criterion \(Q_{i}\) is dependent of the number on feature points $n$ in a patch:
%around a match $m_{i}^{j}$ between some feature $f_{i}$ in image $t$ and $f_{j}$ in image $t+1$, 
%Given a certain number of feature points in a group, \(Q_{i}\) describes how many matches do we expect to be found between them to reliably distinguish between correct matches or mismatches:
%Given our derived probabilities yields:
\begin{equation}
\label{quality:criterion}
Q_{i}=\begin{cases}
    B(n,p_{t}^{cc}),& \text{if}\quad m_{i}^{j}\quad \text{is true}\\
    B(n,p_{f}^{cc}),& \text{if}\quad m_{i}^{j}\quad \text{is false}
  \end{cases}
\end{equation}
%%%\vspace{-1.0em}
\begin{equation}
\label{true}
\mu_{t}=np_{t}^{cc}, \sigma_{t}=\sqrt{np_{t}^{cc}(1- p_{t}^{cc})} \quad \text{if}\quad m_{i}^{j}\quad \text{is true}
\end{equation}
\begin{equation}
\label{false}
\mu_{f}=np_{f}^{cc}, \sigma_{f}=\sqrt{np_{f}^{cc}(1- p_{f}^{cc})} \quad \text{if}\quad m_{i}^{j}\quad \text{is false}.
\end{equation}

\begin{figure}
%%%\vspace{-1.0em}
\centering
\includegraphics[width=0.35\linewidth]{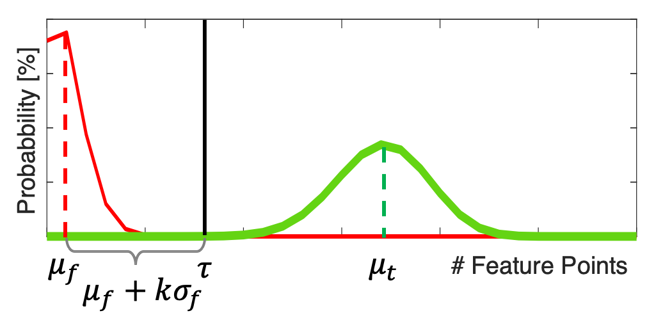}
\caption{Qualitative illustration of the \textit{matching quality criterion} together with the support threshold $\tau$. True and False cases have a wide separation dependent on the number of feature points in the cluster.
%\ben{Description missing. Something like: "The red line shows the probability of wrong assignments p-f-cc and the green line depicts a correct matches p-t-cc for different numbers of feature points." Also the axes are not with numbers, this may be confusing for reviewers. Call the nr. of feature points also "n" as in the text. Is it possible to have a vector graphic here for resolution.}
}
\label{Fig:thresh}
%%%\vspace{-0.8em}
\end{figure}

From a statistical viewpoint, this allows us to formulate a reliable criterion to decide whether or not two groups are correlated and therefore enclose true matches.
%\textbf{Statistical Quality Criteria. }%Our \textit{spatial constraint} determines a true or false feature correspondence based on the number of supporting matches between the associated groups regularized by the \textit{temporal constraint}. 
%With the binomial distribution from above, we can identify a criterion we can identify
The objective is to identify a wide separation between true and false cases. Such a division is given, if one event is at least \(k = 2\) standard deviations $\sigma_{f}$ apart from the mean $\mu_{f}$ (cf. Fig.~\ref{Fig:thresh}). This reduces the probabilities to a simple threshold \(\tau\).

As \(\mu_{f}\) is small (see Eqs.~\eqref{FMP} and ~\eqref{false}) and \(\sigma_{f}\) is mainly dependent on the number of features \(n\) (for $\sigma_{f}$ in Eq.~\eqref{false}, the $p_{f}^{cc}(1- p_{f}^{cc})$ becomes very small), we can write the \textit{support threshold} as:
%which are at least \(k\) standard deviations $\sigma_{f}$ away from the mean number of matches for false correspondences$\mu_{f}$ (cf. Fig.~\ref{Fig:thresh}):
\begin{equation}
\label{threshold}
\tau = \mu_{f} + k\sigma_{f} \approx k\sqrt{n}.
\end{equation}
%These relationships are also depicted in Fig.~\ref{Fig:ProbPlot} where the plotted \textit{support threshold} states how many supporting matches need to be found (given a certain number of features in the group) with cross check between potential group associations in order to consider them true positives.
%The superimposed figures illustrate both cases, for finding enough matches between groups and for discarding the correspondences due to insufficient support.
%\begin{figure}[htbp]
%%%%\vspace{-1.0em}
%\centering
%\includegraphics[width=1.\linewidth]{graphics/fig_prob_illu.pdf}
%\caption{Probabilities for true and false matches of feature points between associated groups with respect %to the number of enclosed feature points in the group.}
%\label{Fig:ProbPlot}
%%%%\vspace{-1.0em}
%\end{figure}
%\ben{I changed the equal sign to approx as we do this here.}
For a given number of features $n$ in a group, we can compute \(\tau\) and compare with the number of other supporting matches between the patches. Is the number of supporting matches higher than \(\tau\), the patches are correlated and the feature matches between them identified as correct.
\section{Experimental Evaluation}\noindent
We quantitatively compare our method against a number of proposed \textbf{classical} matching approaches GMS~\cite{bian2017gms}, SIFT~\cite{Lowe2004}, SURF~\cite{Bay2008}, ORB~\cite{ORB_features}, BD~\cite{lipman2014feature}, BF~\cite{Lin2014}, GAIM~\cite{collins2014analysis}, USC~\cite{raguram2013usac} as well as \textbf{learning} based methods DM~\cite{weinzaepfel2013deepflow} and  LIFT~\cite{Yi2016}. 
We compare on different datasets with small (TUM~\cite{sturm2012benchmark}) and large (Kitti~\cite{geiger2013vision}) baselines as well as scenes with little texture (Cabinet~\cite{sturm2012benchmark}).

Evaluation aspects are based on matching accuracy, robustness and runtime.
To quantify matching accuracy, we evaluate the accuracy of pose estimation from matched features and follow the evaluation protocol of Bian et al.~\cite{bian2017gms} and use their results for comparison on the TUM split.
Pose success ratio is reported as a measure of correctly recovered poses under a certain error threshold.
%in relation to the number of poses. 
The pose is recovered by the estimated essential matrix from feature matches with a RANSAC scheme.
The improved results over the SOTA confirm that our proposed \textit{spatial-temporal} probabilistic model is beneficial in a wide range of textured scenes and different baselines. We observe less convincing results in scenes with limited texture ("Cabinet", see Fig.~\ref{Fig:image_teaser} as example) and explain this as limited ability to form feature groups for such scenes. We justify our assumption by analyzing the average inlier ratio of feature matches of the RANSAC scheme during pose estimation. 
%\todo{Repeatability}
Matching repeatability is analyzed as reprojection error of feature matches in static scenes.
Additional qualitative results are provided as well as an ablation study by disabling parts of the method, thus examining the limitations of our approach.

All experiments are conducted on an Intel Core i7 CPU. We use the publicly available ORB implementation of OpenCV~\cite{opencv_library}.
%for keypoint detection and description with Hamming distance in feature space for NN computation. 
For more details on the maximum number of extracted feature points and parametrization of UFDS please refer to the suppl. material.
%%\vspace{-1.0em}

\begin{figure}[h!]
%%%%\vspace{-1.0em}
\centering
  \includegraphics[width=1.0\linewidth]{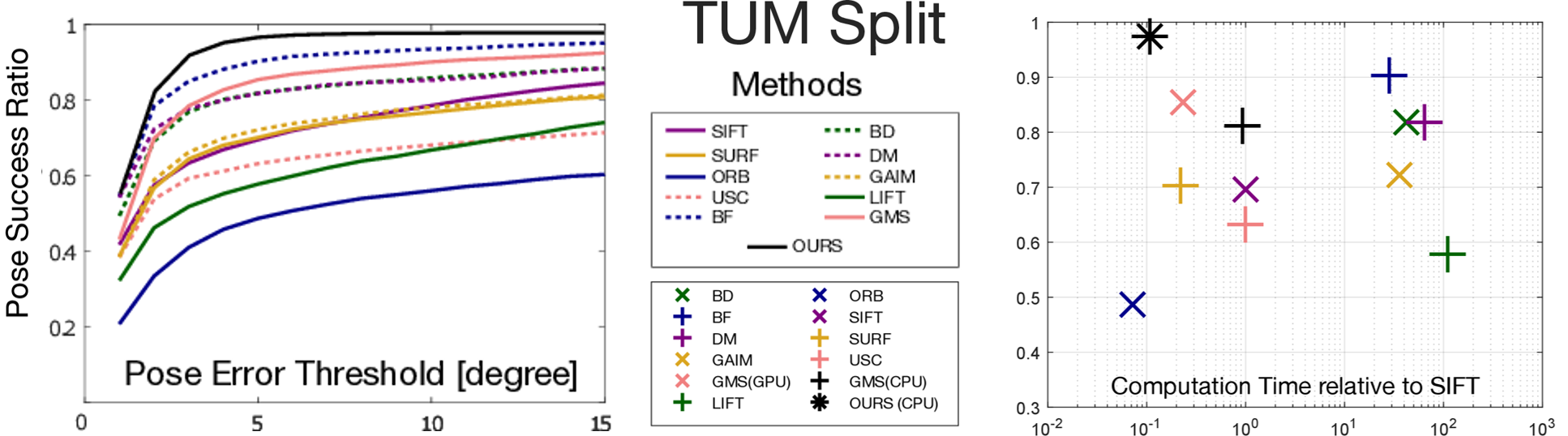}
  \caption{Results on TUM Split~\cite{sturm2012benchmark} with varying scene structure. 
  %The pose success ratio describes the relative amount of correctly estimated poses with different error thresholds. 
  [Left] \textbf{Matching Accuracy} as pose success ratio against pose error threshold.
  [Right] \textbf{Runtime vs. Accuracy} as pose success ratio in relation to computation time (log time scale).
  %The left shows the evaluation on videos with different levels of texture while the right \ben{...}
 }
  \label{split_error}
  %%\vspace{-0.5em}
\end{figure}

\textbf{Matching Accuracy.} 
%The results are shown in Fig.~\ref{Fig:PoseError:ToDo}.
Evidently, \textit{DynaMiTe} outperforms other methods in textured scenarios~\cite{sturm2012benchmark}, as the full potential of our joint formulation of \textit{spatial} and \textit{temporal} constraints can unfold (Fig.~\ref{split_error} [Left]). 

\textbf{Runtime.} We have tested runtime performance on Kitti~\cite{geiger2013vision} and TUM~\cite{sturm2012benchmark}. Our method outperforms SIFT and optical flow (OF)~\cite{Lucas1981a} as baselines and even GMS~\cite{bian2017gms} with GPU acceleration (compare GMS-GPU~\cite{bian2017gms} in Tab.~\ref{table:time}).
\begin{table}
\vspace{-0.5em}
\centering
\scalebox{0.9}{
  \begin{tabular}{lcc|ccc}
   %& OF& SIFT & GMS & GMS* & Ours\\
   & OF~\cite{Lucas1981a}  &  SIFT~\cite{Lowe2004}&  GMS~\cite{bian2017gms}& GMS-GPU & Ours\\ \toprule
Kitti\enskip & 14 & 18 & 3 & 12 & \bf{44}\\
TUM\enskip & 48* & 22 & 4 & 14 & \bf{63}\\
\bottomrule
\end{tabular}}
\caption{
%Comparison with GMS~\cite{bian2017gms}, optical flow (OF)~\cite{Lucas1981a} and SIFT~\cite{Lowe2004} for runtime performance on Kitti~\cite{geiger2013vision} and TUM RGB-D~\cite{sturm2012benchmark} dataset 
\textbf{Runtime} in frames per second (fps). For OF* we report fastest observed fps, as it varies extensively depending on the scene structure and camera displacement.}
\label{table:time}
\vspace{-2.0em}
\end{table}
%%%\vspace{-1.0em}

\textbf{Runtime vs. Accuracy. }
%We test the runtime of \textit{DynaMiTe} against other methods on the TUM-RGB-D dataset~\cite{sturm2012benchmark}. 
For better comparison we evaluate accuracy against runtime (cf. Fig.~\ref{split_error} [Right]). \textit{DynaMiTe} consistently outperforms other methods in terms of runtime vs. success ratio.
%The ideal case would be in the upper left corner, meaning perfect pose success while being fastest.

\begin{figure}[h]
%%%%\vspace{-1.0em}
\centering
  \includegraphics[width=1.0\linewidth]{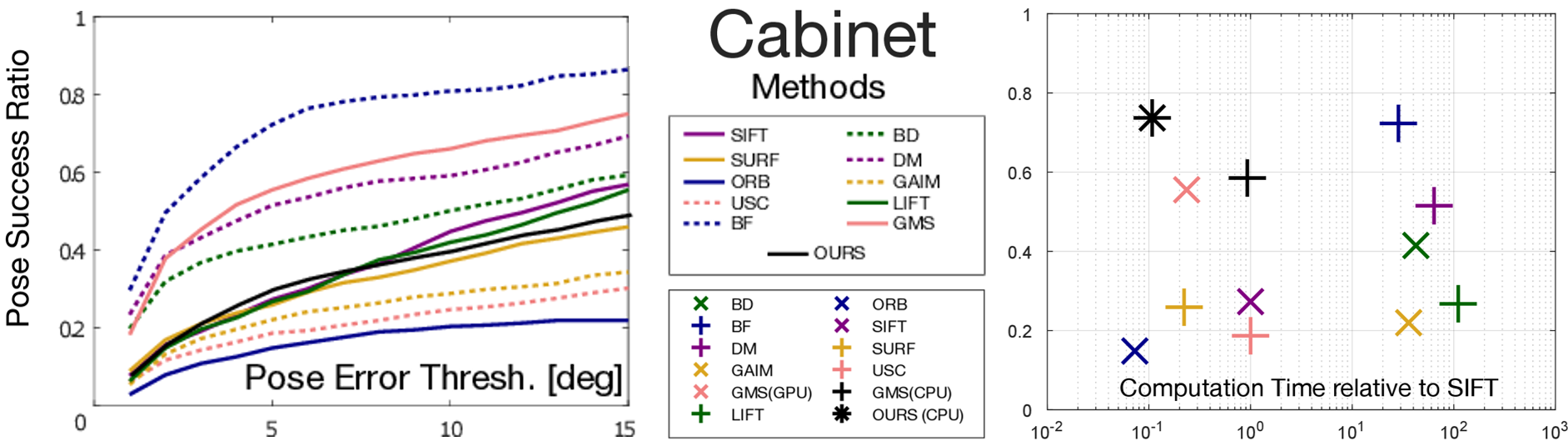}
  \caption{
  Results on low-texture scene "Cabinet"~\cite{sturm2012benchmark}, analogous to Fig.~\ref{split_error}. 
  %Pose success ratio compared against runtime. \textit{DynaMiTe} is the second fastest method (after ORB), while running on CPU only for the complete matching pipeline. 
  %\ben{We need stuff on axes, maybe x can go below, make the colour in the background (not foreground) such that the measures are still the most important part. Why are we here so much better than in the above experiment? Clarify the metrics used (specifically with respect to BF on Cabinet).}
  %Pose success ratio for \textit{DynaMiTe} ranks top. 
  %Runtimes are relative to SIFT as baseline from~\cite{bian2017gms}. Ours on Cabinet performed with baseline up to \(15\deg\) instead of originally up to \(30\deg\) pose error threshold. \ben{Sentence not clear. Sounds cheesy. Say also something about the background colouring}
  }
  \label{cabinet_error}
  %\vspace{-0.5em}
\end{figure}

\textbf{Low-Texture scene. }
For the low-texture scene "Cabinet", tracking 
%\ben{Is it the same as in Fig. 1? Cross reference?})
a large number of group associations throughout the entire sequence is challenging. 
%Hence, many feature matches get lost as they do not comply with our assumptions and 
Only a small number of groups with enough feature points cluster around well defined landmarks.
% Subsequently the success ratio decreases.
\textit{DynaMiTe} still performs on par with other methods, which also have difficulties in this scenario and ranks top in terms of runtime vs. accuracy (compare Fig.~\ref{cabinet_error}).

\textbf{Inlier ratio.}
%To compare the efficiency of the closest methods more easily, 
To analyze the inferior results on "Cabinet",
we add the inliers of RANSAC during camera pose estimation in Tab.~\ref{table:pose_inlierRatio}.
The results reflect our findings, as both pose success and inlier ratio for the textured scenes are superior with \textit{DynaMiTe}, whereas the
Cabinet scene with little structure is challenging.
%This also verifies our assumption that \textit{DynaMiTe} is less powerful in scenes with little structure as establishing many groups of features is challenging.
%%%\vspace{-1.5em}
\begin{table}[!h]
%%\vspace{-0.5em}
    \centering
    \scalebox{0.9}{
    \begin{tabular}{lccc|cc}
          %& OF & SIFT & SIFT* & GMS & Ours\\
          & OF~\cite{Lucas1981a} & SIFT~\cite{Lowe2004} & SIFT* & GMS~\cite{bian2017gms} & Ours\\\toprule
        TUM Split  & 0.58 & 0.16 & 0.54& 0.18 & \bf{0.32}\\
        Cabinet  & 0.50 & 0.20 & 0.61& \bf{0.24} & 0.22\\ 
        Kitti  & 0.37 & 0.11 & 0.64 & 0.85 & \bf{0.87}\\ \bottomrule
    \end{tabular}}
    \caption{Avg. \textbf{inlier ratio} of RANSAC scheme for pose recovery relative to matches. SIFT* includes additional filtering of matches with ratio test.}
\label{table:pose_inlierRatio}
\end{table}
%\vspace{-1.0em}

\textbf{Matching repeatability.} In Tab.~\ref{table:repro} the average match reprojection error for different static scenes from the TILDE webcam dataset~\cite{TILDE} are summarized.
For a perfect match, the norm would be assumed to be \(0\), as the scene and the camera remain static throughout the video. 
This metric can be interpreted as a measure for matching repeatability and the accuracy of the matching scheme as high errors indicate wrong and noisy matches and the inability to robustly handle repetitive patterns.
\textit{DynaMiTe} considerably outperforms SIFT as baseline and reports superior results compared to GMS.
%, especially in Panorama where \textit{DynaMiTe} can recover stable matches in the difficult scene.
\begin{table}
\vspace{-1.0em}
\centering
\scalebox{0.9}{
\begin{tabular}{lcccccc}
 & Chamonix & Courbevone & Frankfurt & Mexico & Panorama &  St. Louis\\
 %& & \cite{bian2017gms} &  \\
 \toprule
 SIFT~\cite{Lowe2004}& 196.03&184.70  & 298.17 & 175.24 & 592.85 & 215.27\\
 GMS~\cite{bian2017gms}&3.48&4.34 & \bf{7.21} & 9.47 &  142.10 &  8.33\\
 Ours&\bf{1.92}&\bf{2.47}&9.45&\bf{6.75}&\bf{2.80} &\bf{3.97} \\ 
\bottomrule
\end{tabular}
}
\caption{Feature \textbf{matching repeatability} test on TILDE dataset~\cite{TILDE} as average \(L_{2}\) reprojection error in pixels.}
\label{table:repro}
\vspace{-1.0em}
\end{table}
%\ben{I would reorder the quantitative experimental section as follows: Matching Accuracy and Runtime, Runtime vs. Acc. on TUM (this is where we are good at, a lot of features...), THEN Cabinet  (make sure to explain cabinet well!, This is the challenging case with no texture) Matching Acc., Runtime, Runtime vs. Acc. THEN Inlier ratio comparison. THEN Matching repeatability on TILDE. THEN failure case.Thus we could start with something really good, then challenging case, then understanding of why.Additionally, the tables are all very small.}

As an additional measure to the evaluation in Tab.~\ref{table:repro}, the error relative to the number of extracted feature points for GMS and our method is analyzed. We calculate the average \(L_{2}\) error normalized per $1000$ features for each sequence and report the average of those as $2.92$ for GMS and $\bf{0.67}$ for \textit{DynaMiTe}, which underlines the favourable efficiency and accuracy of our approach.

\begin{figure}[h]
\centering
  \includegraphics[width=0.8\linewidth]{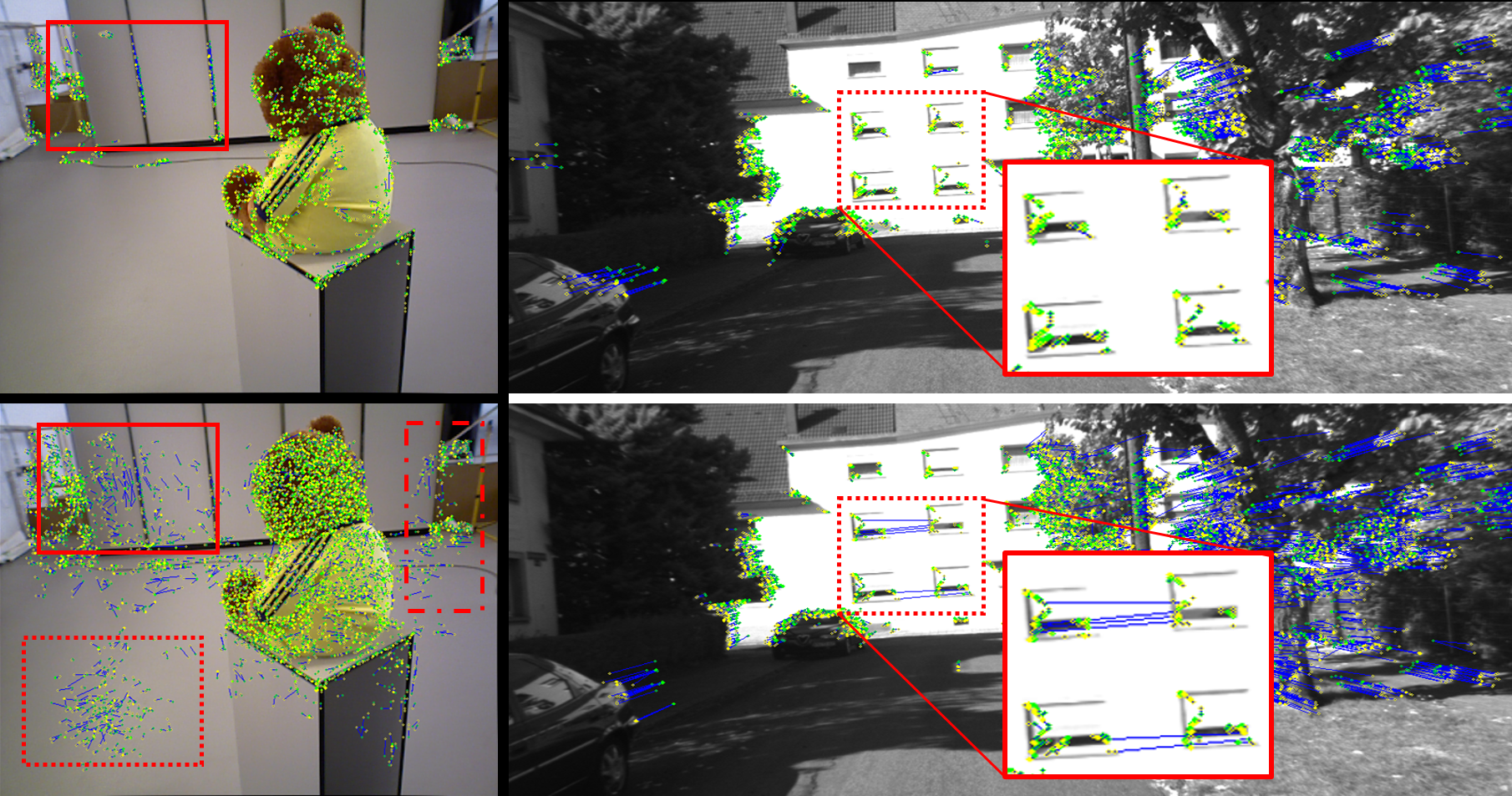}
  \caption{Qualitative robustness comparison on TUM~\cite{sturm2012benchmark} [Left] and Kitti~\cite{geiger2013vision} [Right] dataset. Ours (top) filters noisy and wrong matches in textureless regions and around repetitive patterns.}
  \label{Fig:Video_Robust}
  %%%\vspace{-1.5em}
\end{figure}

\textbf{Qualitative Robustness Evaluation.} We present additional qualitative results on matching robustness in different scenes. Our method filters out noisy, not meaningful matches of the texture-less background.
Furthermore, our proposed cluster grouping and \textit{spatial-temporal} formulation robustly tracks reliable features around landmarks with high image information (e.g. edges and corners of the cabinet, see Fig.~\ref{Fig:image_teaser} and~\ref{Fig:Video_Robust} [Left]).
%; for more examples see the suppl. video.)
%\footnote{The size restriction of the suppl. mat. for the review version only allowed for a low-quality rendering. % In accordance with the review process, A high quality version will be linked with the final version.}
\textit{DynaMiTe} can also handle repetitive patterns in the Kitti sequence, such as the windows on the white building, due to its local clustering algorithm, whereas regular grids such as in GMS fail (see Fig.~\ref{Fig:Video_Robust} [Right]).

\begin{figure}[h]
%%%\vspace{-0.5em}
\centering
  \includegraphics[width=0.8\linewidth]{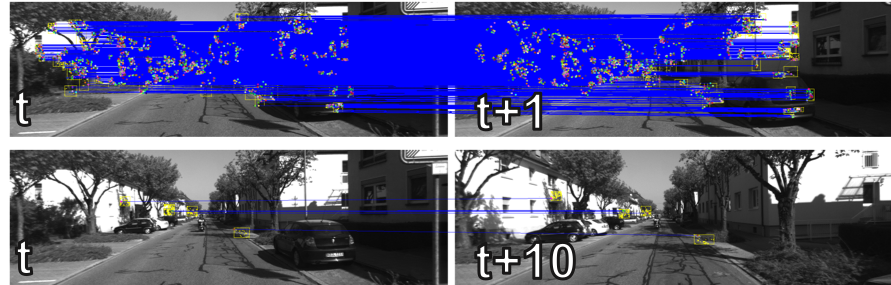}
  \caption{Consecutive frame matching (top) in comparison to limited matching capabilities through multiple frames (bottom) in seq. of~\cite{geiger2013vision}.
  %\ben{Increase size of image.}
  }
  %%%%\vspace{-1.5em}
  \label{Fig:Fail}
  %%%\vspace{-1.5em}
\end{figure}

\textbf{Ablation Study.} The experiments above show applicability on small and wide baseline scenarios (TUM/Kitti). 
%However, for very wide baselines the model is too restrictive. 
%In the the worst case, a single group would essentially span over the majority of the image, resulting in standard feature matching.
Here, we specifically force the algorithm to only keep matches between groups which have been matched throughout the sequence of $10$ consecutive frames and not to establish new group associations between frames. % with wide baselines. 
%Our method normally tries to establish new group associations for unmatched feature points in the image, which has been disabled here.
Due to large inter-frame forward motion, only a few groups in the center of the image are reliably visible throughout all frames. % (e.g. no consistent image information on the side of the image due to fast forward motion).
While our assumptions hold true for consecutive frames, tracking the complete sequence from frame at time step \(t\) to \(t+10\) is problematic as our constraints are violated in this particular setting.
Fig.~\ref{Fig:Fail} illustrates the limitations of our proposed method in this specific case.
%the large inter-frame motion violates the constraints.
%Thereby, the groups cannot be tracked throughout a large number of frames, resulting in a loss of many correct feature matches.
%This shows the limitation of the tightly coupled assumptions, as too few groups would be consistently tracked. 
\textit{DynaMiTe} can still be applied in scenarios with very large baselines, however at the cost of a relaxed constraint for inter-frame motion by increasing the search space for the temporal motion prior.
%\ben{Don't put limitations in conclusion and certainly not the last thing of the paper as we want a positive end. Move it to end of experiments.}
\section{Discussion}\noindent
The reported results clearly show the fundamental trade-off between the ability to correctly match feature points and comply with the runtime constraint for different matching methods. \textit{DynaMiTe} reduces this limitation with its joint formulation, as 
%of spatial and temporal information, as it 
%Our fast clustering scheme enables matching of features between individual dynamically adaptable groups.
it efficiently passes information throughout the sequence, encapsulated in the joint \textit{spatial-temporal} model
This enables very robust feature matching, as well as reduced noise in low-textured scenes, and high framerates without GPU acceleration.
High-confidence noise-free feature matches are beneficial for camera pose estimation, which is what our method focuses on.
The same holds true for reconstruction purposes, one of the various possible application scenarios for \textit{DynaMiTe}.
Generally, our proposed pipeline utilizes solely the information of the feature descriptor and its pixel location in the image, while being agnostic to the underlying descriptor itself. 
Our model achieves robust feature matching even in difficult scenarios and arbitrary inter-frame motion such as scaling and in-plane rotations, as we rely neither on regular grids nor restrictive clustering methods.
%, specifically targeting accurate real-time camera pose tracking.\\
%The input stream for visual odometry and SLAM inherently consists of small baseline samples from consecutive frames similar to the TUM Split in our evaluation part where our method outperforms the state-of-the-art.
%For textureless scenes, our tight spatial-temporal concept is not beneficial, as not enough correct feature matches between descriptive groups can be established, and other methods perform better in terms of pose accuracy. However, these methods cannot cope with the real-time constraint.
%Methods like GMS~\cite{bian2017gms} are favored by the subsequent pose estimation itself with a large set of feature points, whereas \textit{DynaMiTe} keeps only track of consistent and well defined feature points. 

%\ben{This was very defensive (mainly from rebuttals I guess. I tried to make it a little more optimistic.}
%Together with precise camera poses, reconstruction applications benefit due to significantly better initialization. This directly affects the convergence and performance of a BA-based non-linear optimization for 3D reconstruction or camera tracking.\\
\newpage

\clearpage
\bibliographystyle{splncs}
\bibliography{refs_mendeley_dynamite_patrick}

\clearpage
\appendix
\section{Details on 3.3 False Positive Reduction}\noindent
Here, we detail the derivation of Eq.~\eqref{TMP1} from the main paper for better understandability.

\textbf{True Matches. }Let the probability of a feature in \(A\) having its Nearest Neighbor in \(B\) under cross check be \(p_{t}^{cc} = p\left (f_{A}^{B}\mid T, cc\right)\), then it holds:

\begin{align*}
p_{t}^{cc} &= p\left (f_{A}^{B}\mid T,cc  \right )  \\
&= \left(p\left(f_{A}\mid T\right) + p\left(\overline{f_{A}},  f_{A}^{B}\mid T\right)\right) \cdot \left(p\left(f_{B}\mid T\right) + p\left(\overline{f_{B}},f_{B}^{A}\mid T  \right)\right) \\
&= \left(p\left(f_{A}\mid T\right) + p\left(\overline{f_{A}}\mid T\right) \cdot p\left(f_{A}^{B}\mid \overline{f_{A}}, T\right) \right) \\
& \qquad\qquad\qquad\qquad\cdot \left(p\left(f_{B}\mid T\right) + p\left(\overline{f_{B}}\mid T  \right) \cdot p\left(f_{B}^{A}\mid\overline{f_{B}}, T  \right)\right) \\
&\\
& \qquad \rlap{\footnotesize\texttt{as we are independent of $T$ and with} $p\left(f_{A}^{B}\mid \overline{f_{A}}\right) = \frac{n}{N})$,} \\
& \qquad \rlap{\footnotesize\texttt{where \(m\) and \(M\) are equivalent to \(n\) and \(N\):}} \\
&\\
&= \left(p\left(f_{A}\right) + p\left(\overline{f_{A}}\right) \cdot \dfrac{n}{N}\right) \cdot \left(p\left(f_{B}\right) + p\left(\overline{f_{B}} \right) \cdot \dfrac{m}{M}\right) \\
&\\
& \qquad \rlap{\footnotesize\texttt{with $p\left(f_{A}\right) = p\left(f_{B}\right) = t$:}} \\
&\\
&= \left(t + \left(1-t\right) \cdot \dfrac{n}{N}\right) \cdot \left(t + \left(1-t \right) \cdot \dfrac{m}{M}\right) \\
&\\
& \qquad \rlap{\footnotesize\texttt{after binomial expansion:}} \\
&\\
&=t^{2} + 2 \cdot t \cdot \left ( 1-t \right )\dfrac{n}{N} + \left ( 1-t \right )^{2} \dfrac{n}{N} \cdot \dfrac{m}{M}
\end{align*}

\textbf{False Matches.} In analogy for uncorrelated patches:

\begin{align*}
p_{f}^{cc} &= p\left(\overline{f_{A'}},  f_{A'}^{B'}\mid F  \right )\cdot p\left(\overline{f_{B'}},  f_{B'}^{A'}\mid F  \right) \\
&= p\left(\overline{f_{A'}}\mid F  \right)  
\cdot
p\left(\overline{f_{A'}}\mid f_{A'}^{B'}, F  \right)
\cdot 
p\left(\overline{f_{B'}}\mid F  \right)
\cdot
p\left(\overline{f_{B'}} \mid f_{B'}^{A'},  F  \right) \\
&= p\left(\overline{f_{A'}}  \right)  
\cdot
\dfrac{n}{N}
\cdot 
p\left(\overline{f_{B'}}  \right)  \cdot \dfrac{m}{M} \\
&=\left ( 1-t \right )^{2} \dfrac{n}{N} \cdot \dfrac{m}{M}
\end{align*}

%\section{Details on 3.4 Robust Statistics}\noindent

\section{Runtime Analysis}
\begin{figure}[h]
%\vspace{-2.0em}
\centering
  \includegraphics[width=0.8\linewidth]{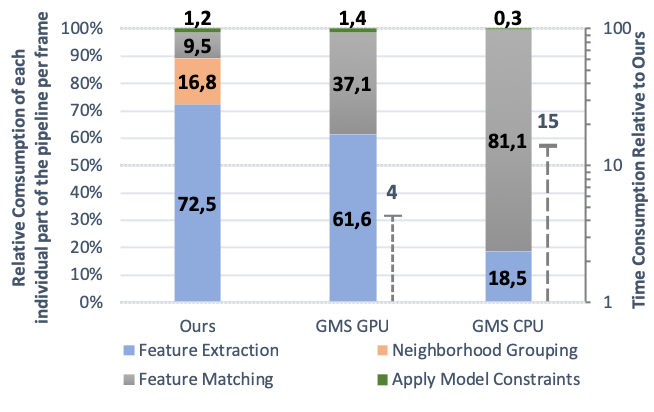}
\caption{Relative runtime comparison. Primary axis shows the relative time consumption per frame for each step of the feature matching pipeline in percentage (numbers are also depicted in the respective bar). Secondary axis (log scale) shows the overall time consumption relative to our proposed method.}
  \label{Fig:Time}
  %\vspace{-1.0em}
\end{figure}

\noindent
Our method achieves realtime performance on CPU for the full pipeline from feature extraction, matching and applying our \textit{spatial} and \textit{temporal constraints}, without any GPU acceleration.
In Fig.~\ref{Fig:Time} the runtime advantage of our method against GMS is clearly visible. GMS is by a factor of 4 slower with GPU acceleration and for CPU-only even by a factor of 15. The matching step, which contributes to a majority of the overall time consumption for GMS and other methods, has now been decreased significantly. The bottleneck for our proposed method is now solely the feature extraction itself.
\section{Parameter Discussion}
%\vspace{-0.5em}
\subsection{Features Points}\noindent
%\vspace{-2.5em}
\paragraph{Extraction. }
We limit the maximum number of extracted feature points in the image. 
Speaking purely from the perspective of estimating camera poses, a small number of feature matches is sufficient. 
However, for our proposed method we assume a certain number of feature points to be detected for forming local feature groups from feature clusters around well defined structures in the image:
\begin{gather}
%\vspace{-1.0em}
\text{Max \#Features} = 7000
\end{gather}
%\vspace{-3.5em}
\paragraph{Descriptor. }
The FAST threshold of ORB is set to $5$ to ensure a high number of detected feature points while not compromising the feature descriptor quality:
\begin{gather}
%\vspace{-1.0em}
\text{FAST threshold} = 5
%\vspace{-1.0em}
\end{gather}
%\vspace{-2.0em}
\subsection{Local Motion Model}\noindent
The parameters for our local motion model are justified by our proposed probabilistic model. Chosen parameters have been used throughout our evaluation, and have therefore been proven to be applicable for different image content and scenarios.
%\vspace{-1.0em}
\paragraph{Group Area.}
We define a maximum size for a local group in pixels:
\begin{gather}
%\vspace{-1.0em}
\text{Group Size} = 30px \times 30px
%\vspace{-1.0em}
\end{gather}
For every feature, the algorithm will find its neighbors within a 30-pixel-by-30-pixel region centering around the feature.
Accompanied with a certain size limit of groups, it enables more nearby features being grouped into a group while keeping the group's size in an appropriate range. This parameter may be adjusted for HR images.
%\vspace{-1.0em}
\paragraph{Group Size.}
Derived from our probabilistic model, we need a minimum number of features per group for applying the statistical criteria.
%(compare Fig.~\ref{Fig:ProbPlot}). 
A maximum number of feature points per group should also be considered. The maximum number is to prevent the group from exceeding expansion, and for very large numbers of features, the quality criterion reaches a saturation stage.
%\begin{figure}[htbp]
%%%\vspace{-1.0em}
%\centering
%\includegraphics[width=1.\linewidth]{graphics/fig_prob_illu.pdf}
%\caption{Probabilities for true and false matches of feature points between associated groups with respect to the number of enclosed feature points in the group.}
%\label{Fig:ProbPlot}
%%%\vspace{-1.0em}
%\end{figure}
%\vspace{-0.5em}
\begin{gather}
\text{Min \#Features in group} = 5\\
\text{Max \#Features in group} = 35
\end{gather}
%\vspace{-1.0em}
%\paragraph{group Density}
%This setting has the same purpose of the maximum size threshold but approaches it in another aspect.
%The problem is: since we are using rectangles to approximate the group's shape, the rectangle may cover a large area while the inside features are very sparse, and in this case the features contained by the group may not represent the underlying regional structure.
%Thus to guarantee a better approximation, we limit the size of the groups.
%$Max area of group = 1/400 of the image size ((width/20) * (height/20))$
%\vspace{-1.0em}

\section{Qualitative Results}
\begin{figure*}[htbp]
\centering
\includegraphics[width=0.99\textwidth]{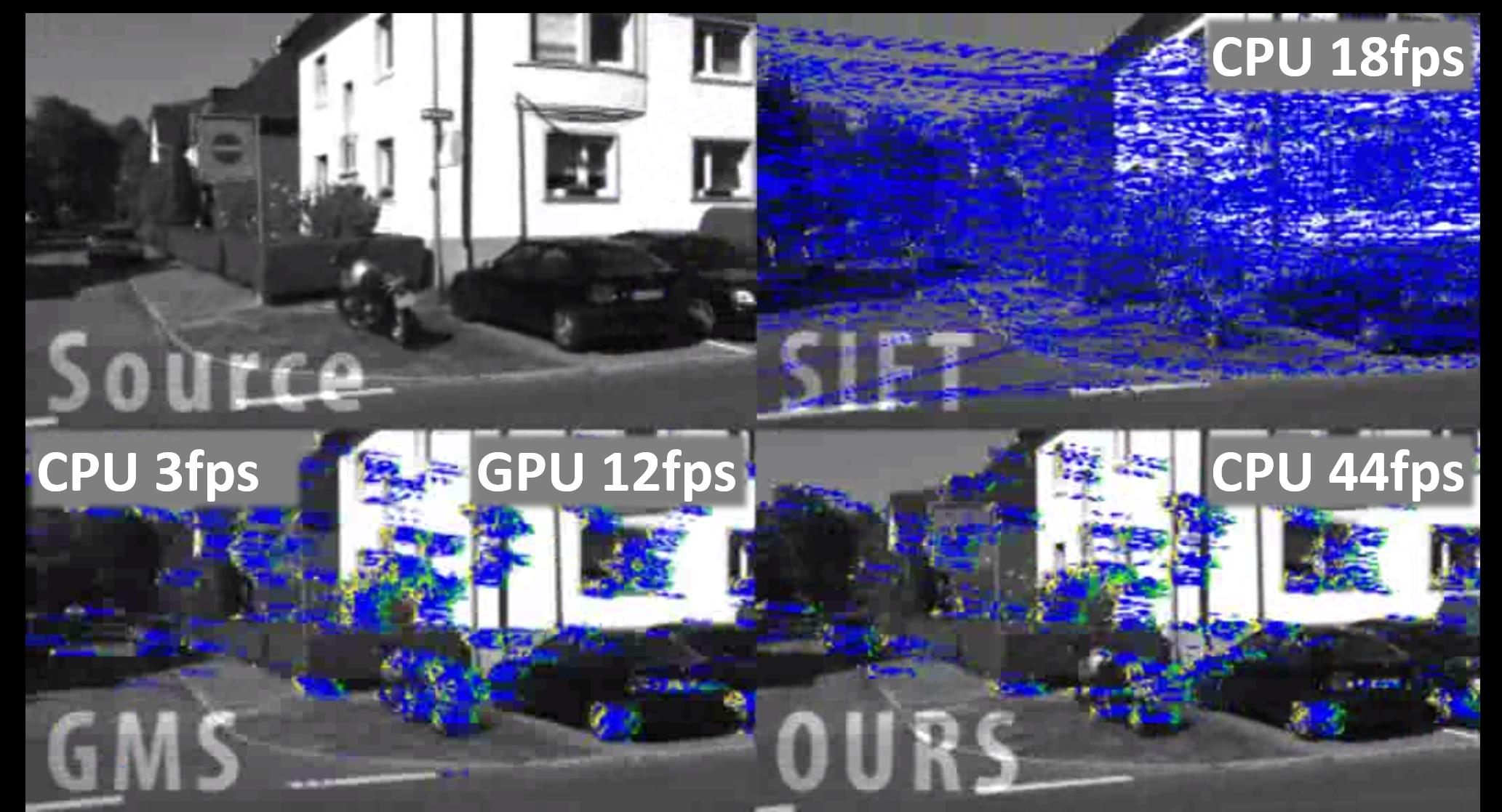}
\caption{Direct comparison of SIFT, GMS, and DynaMiTe (Ours) together with the corresponding runtime on a driving scene from \cite{geiger2013vision}.}
\label{Fig:video_runtime2}
\end{figure*}

\begin{figure*}[htbp]
%%\vspace{-1.0em}
\centering
\includegraphics[width=0.9\linewidth]{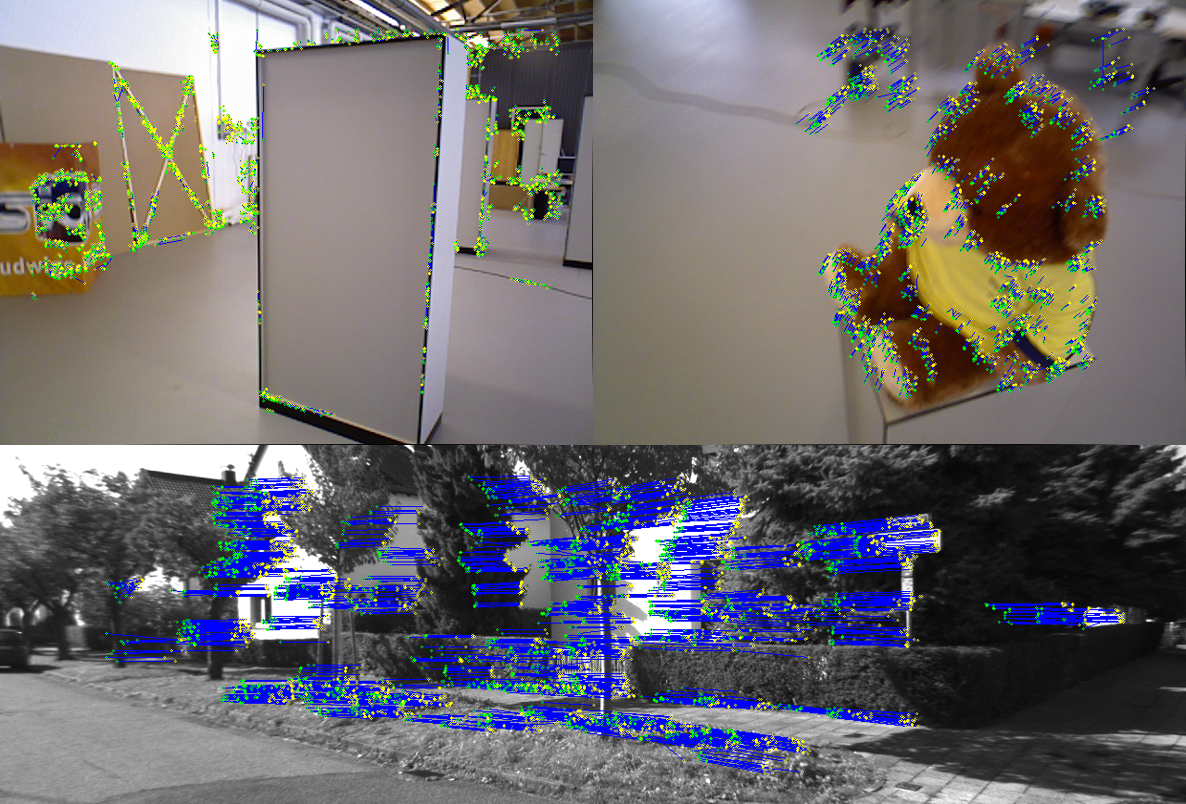}
\caption{Examples of feature point matches for the TUM-RGBD ~\cite{sturm2012benchmark} and Kitti dataset~\cite{geiger2013vision}. Note the challenging scenes with blur (top right) and large rotations (bottom).}
%Epipolar distance errors are indicated below.%\todo{add error plot
%\ruiqitodo{prepare images}
\label{examples}
%%\vspace{-1.0em}
\end{figure*}

\begin{figure}[htbp]
%%%\vspace{-1.0em}
\centering
\includegraphics[width=0.9\linewidth]{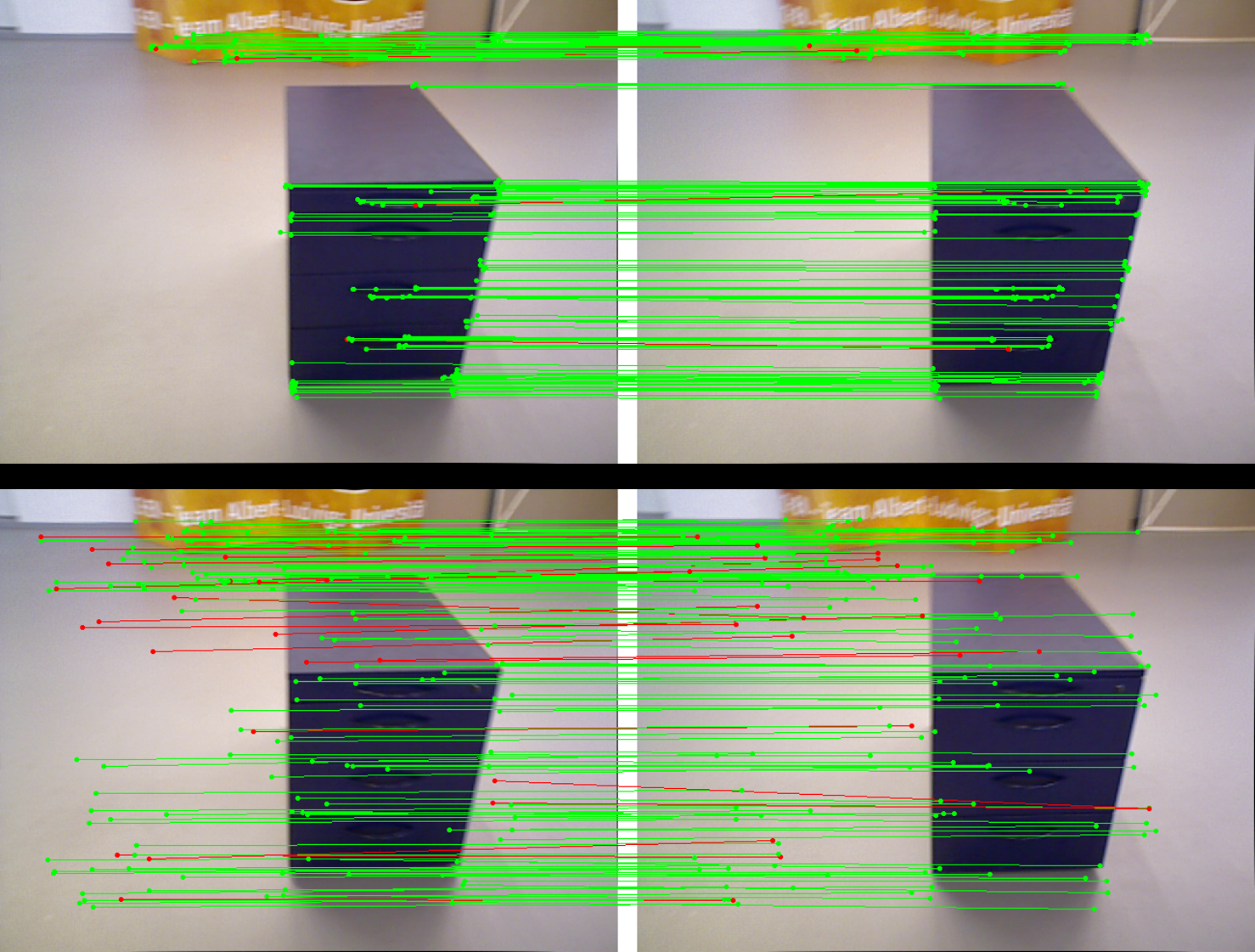}
\caption{Ours (top) reliably tracks only stable feature points as opposed to GMS.}
\label{Fig:QR_1}
%%%\vspace{-1.0em}
\end{figure}

\begin{figure}[htbp]
%%%\vspace{-1.0em}
\centering
\includegraphics[width=0.9\linewidth]{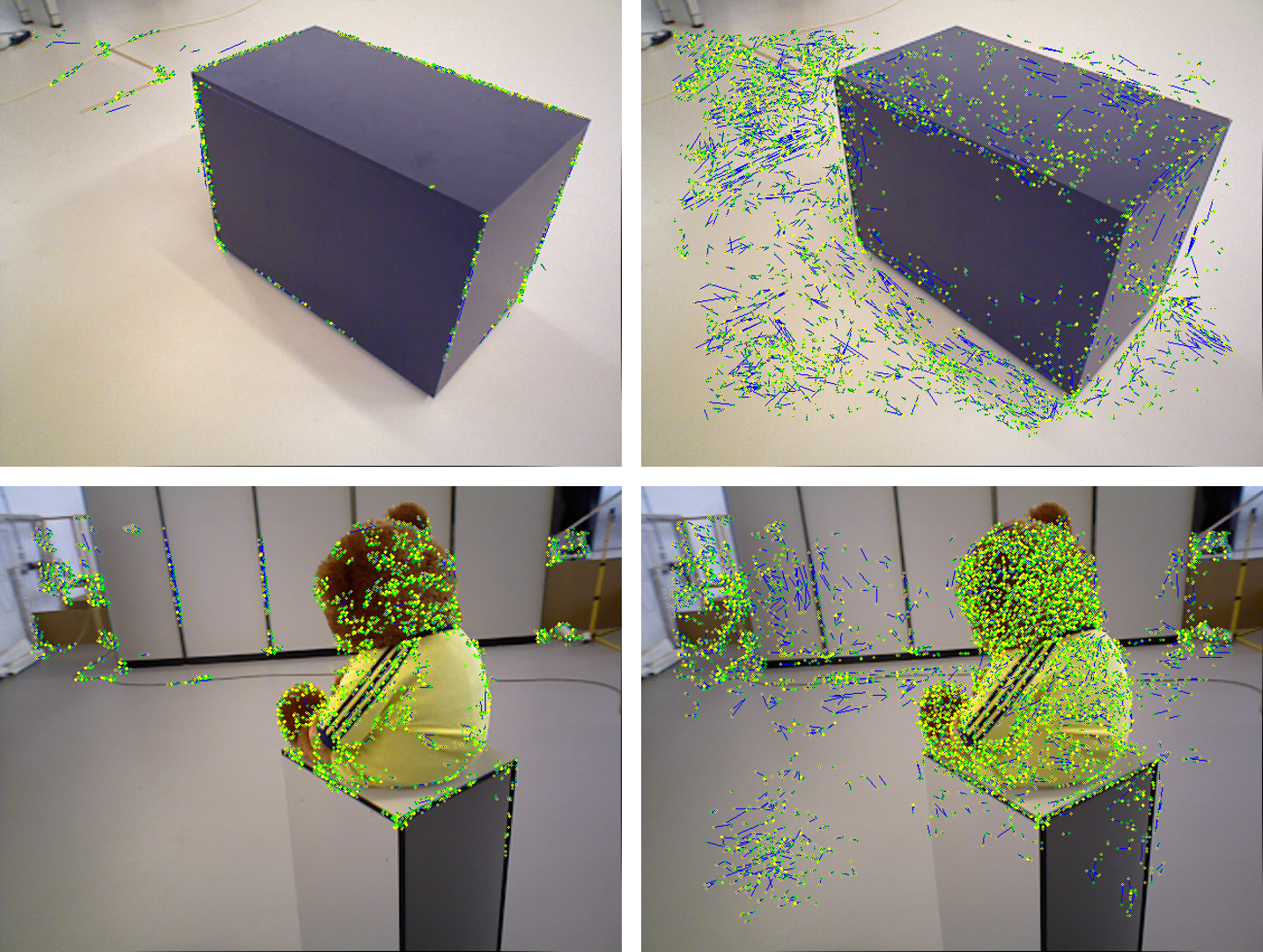}
\caption{Generally ours (left) has reduced noise and less false positive matches in textureless areas compared to GMS.}
\label{Fig:QR_3}
%%%\vspace{-1.0em}
\end{figure}

\noindent
Figures~\ref{Fig:video_runtime2},~\ref{examples},~\ref{Fig:QR_1} and ~\ref{Fig:QR_3} illustrate a few more qualitative results on different datasets. See figure description for more details.

%\begin{figure*}[htbp]
%%%\vspace{-1.0em}
%\centering
%\includegraphics[width=0.9\textwidth]{graphics/QR_2.png}
%\caption{Ours (top) can handle repetitive patterns such as the windows on the white building due to its local clustering algorithm. Regular grids such as in GMS fail.}
%\label{Fig:QR_2}
%%%\vspace{-1.0em}
%\end{figure*}

%\begin{figure*}[htbp]
%\centering
%\includegraphics[width=0.99\textwidth]{graphics/stereoSintel.png}
%\caption{\todo{!!!}}
%\label{Fig:stereoSintel}
%\end{figure*}

\clearpage
\section{Algorithms}\noindent
For a better understanding of our proposed method together with the source code, we provide an overview of the pipeline as pseudo-code.
An overview of the overall pipeline can be found in Algorithm~\ref{DynaMiTe_Algo}.
The grouping algorithm for finding dynamic local feature groups is summarized in Algorithm~\ref{Grouping_Algo}.

\begin{algorithm*}
\KwData{Image \(I_t\); groups \(N_{t-1}\) from \(I_{t-1}\)\ \tcp*{\(N_{t-1}\) from temporal constraint}}
\KwResult{True group Matches \(\in N_t\)}
\(F_t\) = all feature points in \(I_t\)\tcp*{see Algo.~\ref{Grouping_Algo}}
\(N_t\) = GroupingWithUFDS(\(F_{t}\))\;
\(M_t\) = (empty) collection of potentially matched groups\;
\ForAll{\(N^i \in N_t\)}{
    
    \ForAll{\(N^j \in N_{t-1}\)}{
    
        \If{\(N^i\) intersects with \(N^j\)\tcp*{Apply temporal constraint}}{
        \(m^i_j\) = (empty) collection of feature matches per group\;
            \ForAll{\(F_t \in N^i\)}{
                \ForAll{\(F_{t-1} \in N^j\)}{
                        Perform Cross-Check Matching between all \(f_t^k \in F_t\) and all \(f_{t-1}^l \in F_{t-1}\)\;
                            \If{is\_a\_Match}{
                            \(m^i_j\) += Match\;
                            }
                        }
                    }
                
                \(M_t\) += \(m^i_j\)\;
                }
                
        }
    
}

\tcc{Apply spatial constraint}
\ForAll{\(m^i_j \in M_t\)}{
    Compute Score \(S\)\;
    \eIf{\(S > \tau\) }{
        \(m^i_j\) = True group Match\;
        Store Score \(S^i_j\) for group Match \(N^i\) with \(N^j\)
    }{
        Delete \(m^i_j\)\;
    }
}

\tcc{Prepare temporal constraint for next frame}
\ForAll{\(N^i \in N_{t}\)}{
    \ForAll{\(N^j \in N_{t-1}\)}{
        
        Find highest Score \(S\) between groups \(N^i\) and \(N^j\)\;
        Enlarge Search Space for \(N^i\) for next Image \(I_{t+1}\)
        
    }
}

\caption{DynaMiTe Pipeline}
\label{DynaMiTe_Algo}
\end{algorithm*}

\begin{algorithm*}
\KwData{Set of all Features \(F_t\)}
\KwResult{Set of all groups \(N_t\)}
\(Q\) = an auxiliary queue\;
\While{\textbf{not} \(F_t.empty()\)}{
    \eIf{\(Q.empty()\)}{
        Create a new group \(N^k\)\ in \(N_t\)\;
        Pick and then remove a feature \(f_{i}\) from \(F_t\)\;
        \(N^k.add(f_i)\)\;
        \(roi\) = an area centering around \(f_{i}\)\;
        \(Center=f_i.pt\)\;
        \(ptCount=1\)\;
    }{\(f_{i}\) = \(Q.pop()\)\;}
    \tcc{Find current set for \(f_{i}\)}
    
    \If{\(ptCount < MAX\_NUM \&\& N^k.area() < MAX\_AREA\)}{
        \(\{n_i\}\) = all features collected in \(roi\)\;
        \ForAll{\(n_i\)}{
            \(N^k.add(n_i)\) \tcp*{Add \(n_{i}\) to current set of \(f_{i}\)}
            \(F.remove(n_i)\)\;
            \(Q.push(n_i)\)\;
            \(ptCount++\)
        }
    }
    \If{\(Q.empty() \&\& ptCount >= MIN\_NUM\)}{
    \(N_t.add(N^k)\)\;
    }
}
\caption{Grouping Function}
\label{Grouping_Algo}
\end{algorithm*}

\end{document}